\documentclass[final,3p,times,twocolumn]{elsarticle}
\usepackage{amssymb}
\usepackage{amsmath}
\usepackage{url}
\usepackage{subcaption}
\usepackage{amsmath,amsfonts}
\usepackage{algorithmic}
\usepackage{algorithm}
\usepackage{array}
\usepackage[caption=false,font=normalsize,labelfont=sf,textfont=sf]{subfig}
\usepackage{textcomp}
\usepackage{stfloats}
\usepackage{amssymb}
\usepackage{booktabs}
\usepackage{tabularx}
\usepackage{graphbox}
\PassOptionsToPackage{dvipsnames}{xcolor}
\usepackage{colortbl}
\usepackage{multirow}
\usepackage{svg}
\usepackage{comment}

\usepackage{lineno}
\journal{Neural Networks}

\begin{document}

\begin{frontmatter}



\title{Biologically-inspired Semi-supervised Semantic Segmentation for Biomedical Imaging} %% Article title


\author[label1]{Luca Ciampi\corref{cor1}} %% Author name
\ead{luca.ciampi@isti.cnr.it}
\author[label1]{Gabriele Lagani\corref{cor1}}
\ead{gabriele.lagani@isti.cnr.it}
\author[label1]{Giuseppe Amato}
\author[label1]{Fabrizio Falchi}

\cortext[cor1]{Corresponding authors. They contribute equally to this work.}
\affiliation[label1]{organization={ISTI-CNR},%Department and Organization
            %addressline={Via G. Moruzzi }, 
            city={Pisa},
            %postcode={56124}, 
            %state={},
            country={Italy}}

%% Abstract
\begin{abstract}
    We propose a novel bio-inspired semi-supervised learning approach for training downsampling-upsampling semantic segmentation architectures. The first stage does not use backpropagation. Rather, it exploits the Hebbian principle \textit{``fire together, wire together''} as a local learning rule for updating the weights of both convolutional and transpose-convolutional layers, allowing unsupervised discovery of data features.
    In the second stage, the model is fine-tuned with standard backpropagation on a small subset of labeled data.
    We evaluate our methodology through experiments conducted on several widely used biomedical datasets, deeming that this domain is paramount in computer vision and is notably impacted by data scarcity. Results show that our proposed method outperforms SOTA approaches across different levels of label availability. Furthermore, we show that using our unsupervised stage to initialize the SOTA approaches leads to performance improvements. The code to replicate our experiments can be found at \url{https://github.com/ciampluca/hebbian-bootstraping-semi-supervised-medical-imaging}
\end{abstract}



%% Keywords

\end{frontmatter}


%%%%%%%%%%%%%%%%%%%%%%%%%%%%%%%%%%%%%%%%%%%%%%%%%%%%%%%%%%%%%%%%%%%%%%%%%%%%%%%
\section{Introduction}
\label{sec:intro}

Semantic segmentation in biomedical images assigns pixel-level class labels to structures like cells, tumors, and lesions, playing a key role in automating computer-aided diagnoses. Recent data-driven deep-learning approaches using CNNs and Transformers have shown excellent results~\cite{10.1007/978-3-319-24574-4_28,DBLP:journals/corr/ChenPSA17,7298965,7785132,10.1007/978-3-030-87199-4_16,9706678,10.1007/978-3-031-25066-89}. However, these methods demand extensive annotated data for supervised backpropagation-based training, which restricts their large-scale application despite their significance in computer vision~\cite{9625988,Isensee_2020,10.1007/978-3-319-46723-8_49,10.1007/978-3-030-87193-2_4,10.1016/j.patcog.2022.108673,10332179}.

Nevertheless, integration between biological mechanisms and deep learning (BIDL) seems a promising direction, tackling not only the substantial demand for data but also allowing complex neural computation with extreme energy efficiency~\cite{schuman2022, shrestha2022}. 
%the scientific community offered alternative learning rules for artificial networks that do not require error backpropagation and constitute more plausible biologically inspired learning mechanisms~\cite{marblestone2016,gerstner,survey_snn,richards2019}. Integration between biological mechanisms and deep learning (BIDL) seems a promising direction, tackling not only the substantial demand for data but also allowing complex neural computation with extreme energy efficiency~\cite{schuman2022, shrestha2022}. 
In particular, the so-called Hebbian principle represents a simple local learning rule that closely mimics the synaptic adaptations of brain mechanisms~\cite{haykin,gerstner,survey_plasticity}, offering an appealing and still relatively unexplored solution for extracting data features without relying on supervised backpropagation and on the availability of large labeled datasets. Different Hebbian learning strategies involving Soft-Winner-Takes-All (SWTA)~\cite{krotov2019, moraitis2021,lagani2022a}, or Principal Component Analysis (HPCA)~\cite{pehlevan2015a, bahroun2017, lagani2021b}, allow a group of neurons to discover unsupervised features from a set of data such as clusters or principal components, and can be used to fulfill unsupervised and semi-supervised learning strategies coming to the rescue to mitigate data scarcity issues. 

\begin{figure*}[t]
    \centering
    \includegraphics[width=0.8\textwidth]{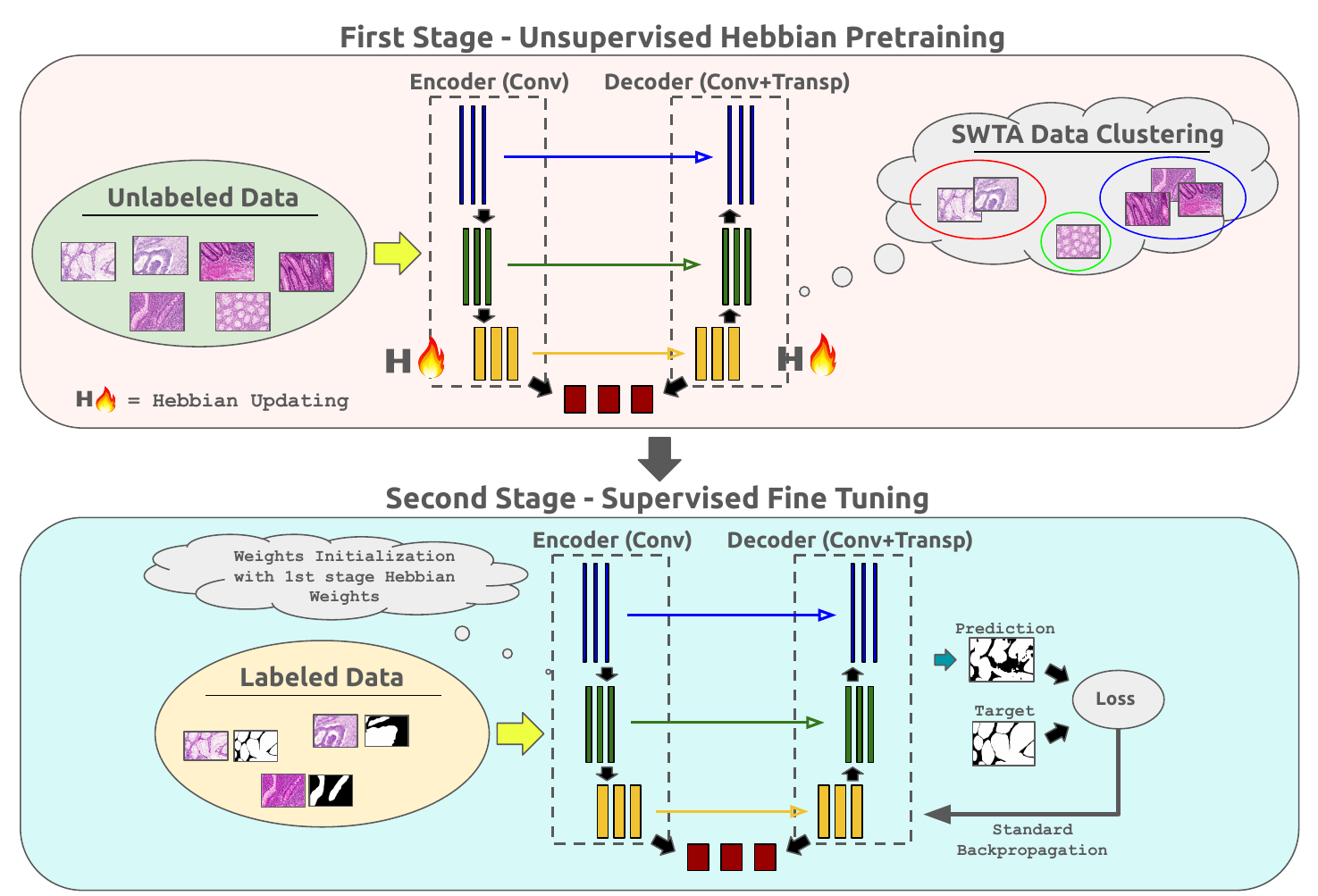}
    \caption{
    Our semi-supervised, bio-inspired approach for semantic segmentation architectures with downsampling and upsampling paths. The method unfolds in two stages. First, we employ Hebbian learning for unsupervised pre-training on a large set of unlabeled data, enabling the model to autonomously identify features like cluster centroids. Notably, we derive new Hebbian learning rules for the transpose-convolution layers in the upsampling path. In the second stage, we apply supervised backpropagation to fine-tune the model using a small labeled dataset.
    }
    \label{fig:teaser}
\end{figure*}

In this work, we introduce a two-stage semi-supervised biologically-inspired learning approach for semantic segmentation in biomedical images (Fig.~\ref{fig:teaser}). In the first stage, we exploit a set of unlabeled data to train downsampling-upsampling semantic segmentation architectures in an unsupervised fashion using the Hebbian learning principle implemented with different strategies. In the second step, we grab these weights to initialize the model, and we fine-tune it through standard backpropagation supervised training on a few labeled data samples. Specifically, during the unsupervised round, we employ Hebbian learning rules in standard convolutional layers in the downsampling path and, for the first time, we formulate novel Hebbian learning strategies for transpose-convolutional (T-Conv) layers 
%needed 
exploited for upsampling the feature maps.

We perform an experimental assessment of our methodology over three public datasets widely used in the context of medical 2D image segmentation~\cite{9625988,10.1007/978-3-030-87193-2_4,10.1016/j.patcog.2022.108673,10332179} showcasing different imaging modalities and segmentation tasks, i.e., GlaS~\cite{SIRINUKUNWATTANA2017489} for colorectal cancer segmentation in Hematoxylin and Eosin (H\&E) stained histological images, PH2~\cite{6610779} for skin lesion segmentation in dermoscopic images, and HMEPS~\cite{raffaele_mazziotti_2021_4488164} for eyes pupil segmentation in grayscale images. Moreover, we also conduct an analysis using the LA~\cite{XIONG2021101832} dataset for left atrial segmentation in MRIs, thus extending our method to volumetric images. We compare our approach against several SOTA single-stage semi-supervised approaches relying on pseudo-labeling and consistency training~\cite{8954439,10.1007/978-3-030-32245-8_67,9157032,9577639,LUO2022102517,Luo_2021}, as well as some other two-stage pipelines exploiting Variational Auto-Encoders (VAEs)~\cite{DBLP:journals/corr/KingmaW13} 
%Denoising Diffusion Probabilistic Models (DDPMs)~\cite{10.5555/3495724.3496298}, 
and Self-Supervised Learning (SSL)~\cite{10.1007/978-3-030-58526-6_45} for the unsupervised step. The outcomes demonstrate that our approach can substantially improve performance considering various label scarcity conditions. 
Furthermore, we show that using our unsupervised Hebbian stage to initialize the SOTA single-stage approaches results in performance improvements.

Concretely, our contributions are the following:
\begin{itemize}
    \item We propose a novel semi-supervised two-stage pipeline for semantic segmentation, where a first unsupervised stage relying on the bio-inspired Hebbian principle is followed by a second supervised step that fine-tunes the model on a few labeled data samples. 
    \item We validate our methodology on several public biomedical imaging benchmarks, demonstrating that our technique can substantially improve performance compared to SOTA methods across different degrees of label availability. 
    \item We conduct a further experimental analysis by initializing existing SOTA semi-supervised approaches with our Hebbian unsupervised pre-training, showing performance improvements.
\end{itemize}

We organize the rest of the paper as follows. In Sec.~\ref{sec:related_works}, we review related work. Sec.~\ref{sec:background} provides a brief background on Hebbian learning. In Sec.~\ref{sec:method}, we describe our methodology, while Sec.~\ref{sec:experiments} presents the experiments, discusses the results, and includes ablation studies to validate our approach. Finally, Sec.~\ref{sec:conclusion} concludes the paper. Additionally,~\ref{appendix:background} and~\ref{appendix:implementation} provide further details on the background of Hebbian learning and the implementation, respectively.

%%%%%%%%%%%%%%%%%%%%%%%%%%%%%%%%%%%%%%%%%%%%%%%%%%%%%%%%%%%%%%%%%%%%%%%%%%%%%%%
\section{Related Works}
\label{sec:related_works}

\subsection{Deep Learning Bio-inspired Approaches}
\noindent Recent Deep Learning (DL) models have achieved remarkable success across various tasks~\cite{devlin2019, CIAMPI2022102500, dosovitskiy2020}. However, key challenges remain, including the substantial demand for data~\cite{roh2019} and energy~\cite{badar2021}. In contrast, biological intelligence appears to overcome these limitations~\cite{lake2020, javed2010}, suggesting that a tighter integration between biological mechanisms and DL (Bio-Inspired DL - BIDL) could be a promising direction~\cite{hassabis2017, lake2017}. A significant area within BIDL is Spiking Neural Networks (SNNs)~\cite{wu2019, lee2020, zhou2021, goltz2021}, which model neural computation in a way that more closely resembles biological neurons. In particular, neuromorphic hardware implementations based on SNNs offer complex neural computation with exceptional energy efficiency~\cite{wang2022, schuman2022, shrestha2022}. Another crucial aspect of BIDL pertains to training algorithms. Conventional DL models rely on backpropagation, which is widely considered biologically implausible~\cite{oreilly}. Consequently, alternative learning strategies inspired by biological synaptic plasticity have gained traction. Specifically, Hebbian learning~\cite{haykin, gerstner, journe2023, lagani2021b, lagani2022a} provides a biologically plausible alternative for learning in DL systems~\cite{bahroun2017, krotov2019, moraitis2021, journe2023, lagani2022a, lagani2022c, hebbian_eccv_workshop}. 

\subsection{Semi-supervised Semantic Segmentation}
\noindent The widespread adoption of deep learning has led to the extensive utilization of CNNs and Transformers in the realm of semantic segmentation, such as FCN~\cite{7298965}, SegNet~\cite{7803544}, and DeepLabV3~\cite{DBLP:journals/corr/ChenPSA17}. Concerning biomedical images, UNet-like architectures have emerged to be the most efficient and performing ones~\cite{10.1007/978-3-319-46723-8_49,7785132,9706678,10.1007/978-3-031-25066-89,Isensee_2020}. To address the primary limitation of these approaches -- namely, the limited availability of labeled data -- semi-supervised techniques offer a promising solution. This paradigm aims to extract knowledge from a vast pool of unlabeled data in combination with supervised learning on a few labeled samples~\cite{bengio2007,larochelle2009}. 

\paragraph{Single-stage approaches} Dominant semi-supervised strategies include single-stage approaches categorized as (i) pseudo-labeling, where the model computes an auxiliary loss generating pseudo-targets associated with the unlabeled data, and (ii) consistency training, where the model is fed with different perturbed versions of a given input and is enforced to generate similar predictions associated to them, constructing an additional loss satisfying this consistency criterion ~\cite{8954439,10.1007/978-3-030-32245-8_67,9157032,9577639,LUO2022102517,Luo_2021}.
Entropy Minimization (EM)~\cite{8954439} is a pseudo-labeling approach that uses predictions of the network itself as pseudo-labels for unlabeled samples, 
%minimizing the cross-entropy between outputs and pseudo-labels. This is equivalent to
minimizing the entropy of the output probability distribution.
%and thus reinforcing the network's confidence in a given prediction. 
Uncertainty-Aware Mean Teacher (UAMT)~\cite{10.1007/978-3-030-32245-8_67} follows a student-teacher architecture in which the student model learns progressively from the teacher model’s reliable predictions. The teacher model not only generates target outputs but also assesses the uncertainty of each prediction using Monte Carlo sampling.
%Mean Teacher (MT)~\cite{10.5555/3294771.3294885} is another notable pseudo-labeling technique. Here, a teacher model is derived from a trainee model, \ie, from the student, and is in charge of computing the pseudo-label for unlabeled samples; specifically, the teacher and the student have the same architecture, but the teacher's weights are computed over the training iterations as an exponential moving average of the student's weights. 
Cross-Consistency Training (CCT)~\cite{9157032} is a consistency-based method that generates different predictions by applying various perturbations to the latent representation generated by the model from a given unlabeled input. A loss term encourages the model to align predictions from different perturbations. A more recent consistency-based approach is Uncertainty Rectified Pyramid Consistency (URPC)~\cite{LUO2022102517} which learns from unlabeled data by minimizing the discrepancy between a set of segmentation predictions at different scales.
%A more recent approach, again belonging to consistency training, has been proposed in~\cite{10376766}. In this work, the authors introduced a Low-High Frequency Consistency (LHFC) criterion: an unlabeled given input is elaborated through a wavelet filter in order to separate the high-frequency wavelet components from the low-frequency ones, thus generating two different input variants. Both are fed to the network for processing, and a consistency criterion aligns the resulting predictions. 
Conversely, Cross Pseudo Supervision (CPS)~\cite{9577639} enforces consistency between two segmentation networks with different initializations for the same input image. Each network generates a pseudo-one-hot label map, which is then used to supervise the other network through a standard cross-entropy loss. Instead, the authors in~\cite{Luo_2021} proposed a Dual-Task Consistency (DTC) framework that jointly predicts a pixel-wise segmentation map and a geometry-aware level set representation of the target.

\paragraph{Two-stage approaches} An alternative semi-supervised approach involves an initial unsupervised training phase on unlabeled data, followed by fine-tuning on labeled samples~\cite{bengio2007,larochelle2009,kingma2014b,zhang2016}. Techniques commonly used in the unsupervised phase include Variational Autoencoders (VAEs)~\cite{DBLP:journals/corr/KingmaW13} and Self-Supervised Learning (SSL)~\cite{10.1007/978-3-030-58526-6_45}. While Hebbian learning-based pretraining has previously been applied to image classification~\cite{lagani2021b}, the use of Hebbian rules for weight updates in T-Conv layers 
%-- enabling unsupervised pretraining in semantic segmentation architectures -- 
has been largely unexplored and was only introduced in a preliminary form in our previous paper~\cite{hebbian_eccv_workshop}. In this paper, we extend~\cite{hebbian_eccv_workshop} by (i) formulating more advanced Hebbian learning rules specifically designed for T-Conv layers; (ii) conducting experiments on additional datasets and comparing our approach against a broader range of state-of-the-art methods; (iii) leveraging our unsupervised stage to initialize SOTA approaches, demonstrating performance improvements; and (iv) providing extensive ablation studies to validate our methodology.

\begin{figure}
    \centering
   \includegraphics[width=0.9\linewidth]{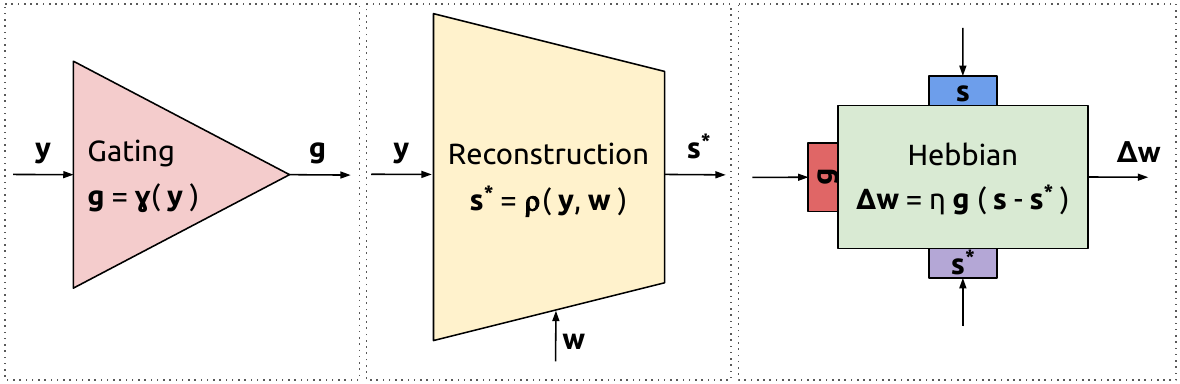}
    \caption{Building blocks for Hebbian learning. Hebbian updates are computed from the difference between a target signal $\mathbf{s}$ and a reconstructed signal $\mathbf{s*}$, and a gating signal $\mathbf{g}$ which modulates the update steps $\mathbf{g}$. Gating and reconstruction signals are, in turn, derived from outputs and weights through the respective blocks.
    }
    \label{fig:hebb_T-Conv_a}
\end{figure}

%%%%%%%%%%%%%%%%%%%%%%%%%%%%%%%%%%%%%%%%%%%%%%%%%%%%%%%%%%%%%%%%%%%%%%%%%%%%%%%
\section{Background}
\label{sec:background}

\noindent This section provides a brief background about Hebbian 
%biologically inspired 
learning rules for training artificial networks. 
More details can be found in~\ref{appendix:background}.

Neuroscientists formulated the plasticity occurring in synapses connecting different brain neurons proposing the Hebbian principle \textit{``fire together, wire together''}.
Essentially, it is grounded on the biological observation that neurons tend to strengthen their connections with other neurons when their activities are correlated, or to weaken their couplings otherwise~\cite{gerstner}.
Mathematically, this behavior can be modeled by a simple learning rule implemented as follows~\cite{haykin}:

\begin{equation} 
    \label{eq:hebbian_vanilla}
    \Delta w_{i, j} = \eta \, y_j \, ( x_i - w_{i, j} ),
\end{equation}

\noindent where $x_i$ is the input stimulus delivered from neuron $i$ to neuron $j$, $y_j$ is the output of neuron $j$, $w_{i, j}$ are the weights connecting a neuron $i$ and a neuron $j$, $\Delta w_{i, j}$ is the weight update computed by the learning rule, and $\eta$ is the learning rate. The intuition behind this learning rule is that if a neuron is exposed to a set of inputs, the weight vector moves toward the centroid of the cluster formed by those inputs and the neuron becomes a detector of such a pattern~\cite{grossberg1976a,rumelhart1985,haykin,lagani2022a}.
Different formulations of this rule led to the Soft-Winner-Takes-All (SWTA) and the Hebbian Principal Component Analysis (HPCA) techniques.

\paragraph{Soft-Winner-Takes-All (SWTA)}
The authors of~\cite{grossberg1976a,rumelhart1985} proposed the idea of \textit{competitive learning}, to allow a set of neurons to specialize on different patterns: this is achieved by assigning higher weight updates to those neurons that best match the current input. Specifically, recent work~\cite{lagani2021b,moraitis2021} demonstrated the effectiveness of Soft-Winner-Takes-All (SWTA) learning, i.e., a form of soft competition where neurons take update steps proportional to the softmax of their outputs:

\begin{equation} 
    \label{eq:swta}
    \Delta w_{i, j} = \eta \, \mathrm{softmax}(y_1, y_2, ...)_j \, ( x_i - w_{i, j} ) .
\end{equation}

\noindent where $\mathrm{softmax}(y_1, y_2, ...)_j = \frac{e^{y_j/t}}{\sum_k e^{y_k/t}}$ and $t$ is the temperature hyperparameter to tune the sharpness of the softmax operation ($t \rightarrow 0$ is equivalent to single WTA competition, while $t \rightarrow \inf$ means no competition). 
%The softmax operation can be computed using a temperature hyperparameter $T$, whose best value depends on the specific task and should be empirically optimized.

\paragraph{Hebbian Principal Component Analysis (HPCA)}
The authors of~\cite{sanger1989,karhunen1995,becker1996a,pehlevan2015a} modeled lateral connectivity patterns observed in the brain~\cite{luo2021, pehlevan2015a} and, from this model, they derived a new learning rule where contributions from neighboring neurons appear in the weight update of each neuron:

\begin{equation} 
    \label{eq:hpca}
    \Delta w_{i, j} = \eta \, y_j \, ( x_i - \sum_{k=1}^j y_k w_{i, k} ) .
\end{equation}

\noindent This formulation can be shown to be equivalent to Principal Component Analysis (PCA)~\cite{sanger1989,karhunen1995,becker1996a}: by following this learning rule, the weight vectors of different neurons align towards the directions of the principal components of the data, in an efficient and biologically plausible fashion.

%%%%%%%%%%%%%%%%%%%%%%%%%%%%%%%%%%%%%%%%%%%%%%%%%%%%%%%%%%%%%%%%%%%%%%%%%%%%%%%
\section{Method} % Hebbian learning for Semantic Segmentation
\label{sec:method}

\begin{figure*}[!t]
    \centering
    \begin{subfigure}[t]{0.29\textwidth}
        \center\includegraphics[width=0.9\textwidth]{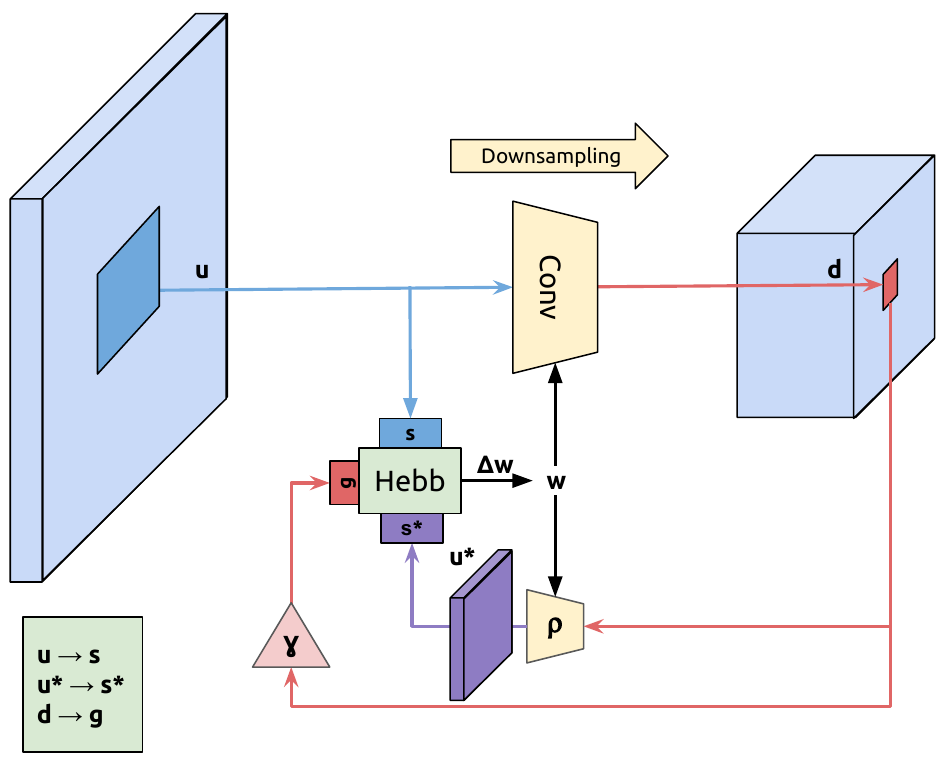}
        \caption{\scriptsize Hebbian update computation in a standard convolutional layer, mapping an upsampled feature map into a downsampled feature map: a patch from the upsampled map is the target signal, while the gate and reconstruction signals are computed from a downsampled map patch through the respective blocks.
        }
        \label{fig:hebb_T-Conv_a_b}
    \end{subfigure}
    ~
    \begin{subfigure}[t]{0.29\textwidth}
        \center\includegraphics[width=0.9\textwidth]{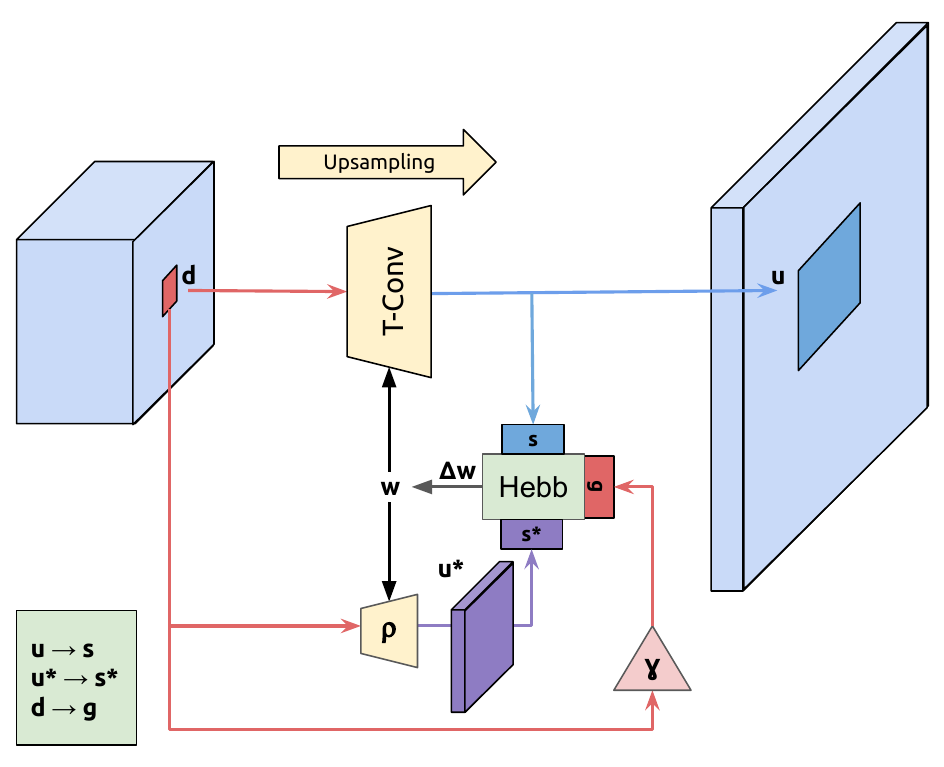}
        \caption{\scriptsize Naive extension of Hebbian update computation to T-Conv layers: as before, we can use a patch from the upsampled map as the target signal, while the gate and reconstruction signals are computed from a downsampled map patch through the respective blocks.}
        \label{fig:hebb_T-Conv_a_c}
    \end{subfigure}
    ~
    \begin{subfigure}[t]{0.29\textwidth}
        \center\includegraphics[width=0.9\textwidth]{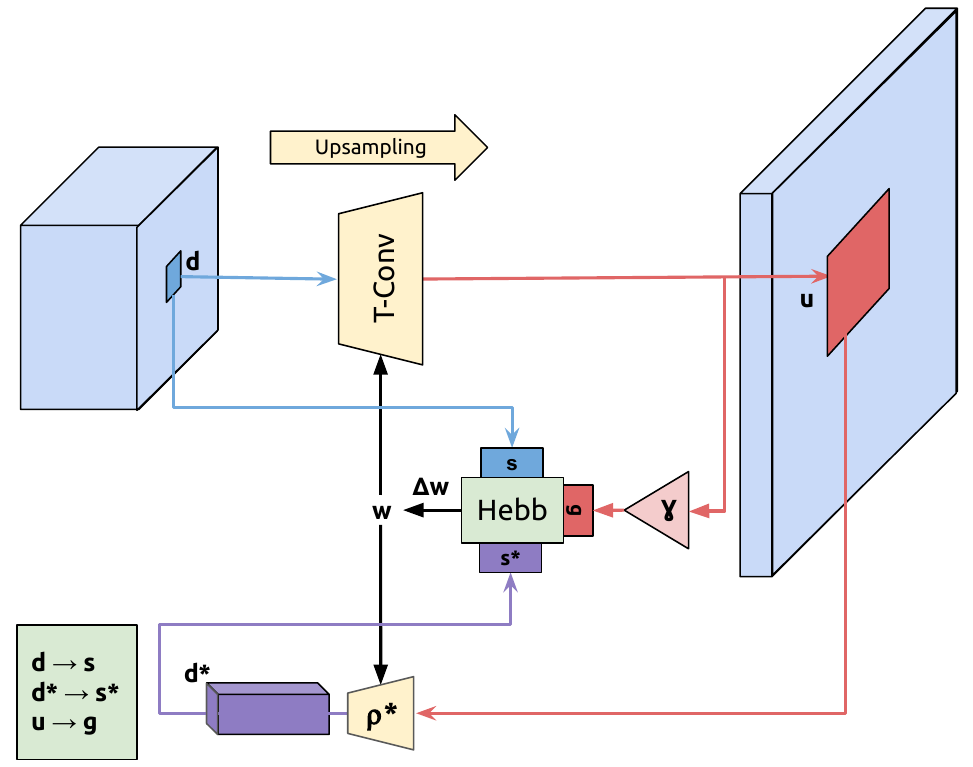}
        \caption{\scriptsize Our formulation of Hebbian update computation to T-Conv layers: a patch from the downsampled map is used as the target signal, while the gate and reconstruction signals are computed from an upsampled map patch through a gating block and a specifically designed reconstruction block.}
        \label{fig:hebb_T-Conv_a_d}
    \end{subfigure}
    \caption{Hebbian learning in convolutional and transpose-convolutional layers.}
    \label{fig:hebb_T-Conv}
\end{figure*}

\subsection{Problem Formulation}
\label{sec:sec:method_semi_supervised}

\noindent We assume to have a dataset $\mathcal{X} = \mathcal{X}_L \cup \mathcal{X}_U$, where $\mathcal{X}_L = \{(x_i, y_i)\}_{i=1}^{N_L}$ is the labeled subset including images $x_i$ and corresponding pixel-wise labels $y_i$, and $\mathcal{X}_U = \{(x_j)\}_{j=1}^{N_U}$ is the unlabeled subset containing only images $x_j$. 
%Let us consider a training set $\mathcal{D}$ and two subsets of it: (i) a \textit{labeled set} $\mathcal{D}_L$ for which label information is known, and (ii) an \textit{unlabeled set} $\mathcal{D}_U = \mathcal{D} \setminus \mathcal{D}_L$ containing the remaining elements. 
We also assume to be in typical semi-supervised settings, i.e., $N_L \ll N_U$ and $\mathcal{X}_L \cap \mathcal{X}_U = \emptyset$. Furthermore, we define an $r \, \%$-\textit{regime} as $\frac{r}{100}|\mathcal{X}_{L}|$, outlining several levels of label availability from $\mathcal{X}_L$ for the supervised training. We formulate a semi-supervised learning pipeline for semantic segmentation based on two steps (Fig.~\ref{fig:teaser}). In the first stage, unsupervised pre-training is performed using Hebbian learning algorithms on $\mathcal{X}_U$. Then, in the second stage, we fine-tune the model using standard backpropagation on $\mathcal{X}_L$ at different regimes $r$.

\subsection{Hebbian Learning for T-Conv Layers}
\label{sec:sec:method_hebbian_tconv}
\noindent Neural networks for semantic segmentation are usually characterized by encoder-decoder architectures, where a downsampling path responsible for computing image features is followed by an upsampling path that restores the spatial dimensions to make predictions pixel-wise~\cite{7298965,7803544,DBLP:journals/corr/ChenPSA17}. Specifically, UNet-like architectures~\cite{10.1007/978-3-319-24574-4_28} have emerged to be the most efficient and performing models for semantic segmentation in biomedical images~\cite{10.1007/978-3-319-46723-8_49,7785132,9706678,10.1007/978-3-031-25066-89,Isensee_2020}. Here, convolutional layers make up the encoding stage, while the decoding step usually relies on T-Conv layers~\cite{7298965} which are in general preferable compared to the combination of convolutional layers and interpolation~\cite{DBLP:journals/corr/DumoulinV16}. 
As shown in Sec.~\ref{sec:background}, Hebbian learning strategies enable the unsupervised discovery of data features such as cluster centroids, and they have successfully been adopted in convolutional and fully-connected layers~\cite{lagani2022c, lagani2024a,moraitis2021,journe2022}. However, a Hebbian theory for T-Conv layers is lacking, (it was only introduced in a preliminary form in our previous paper~\cite{hebbian_eccv_workshop}).
%, making the adoption of such methodologies impossible for semantic segmentation architecture. 
In this section, we illustrate our contributions to fill this gap.

\paragraph{Building blocks of Hebbian learning} For convenience, we decompose the Hebbian learning rules previously described in Eq.~\ref{eq:swta} and Eq.~\ref{eq:hpca} into three basic building blocks, depicted in Fig.~\ref{fig:hebb_T-Conv_a}. 
The first block is a \textit{gating} function $\gamma(\mathbf{y})$, which provides a factor to modulate the size of the update steps based on the neurons' outputs. It corresponds to $\gamma(\mathbf{y}) = \mathrm{softmax}(\mathbf{y})$ (the softmax function) for SWTA, and $\gamma(\mathbf{y}) = \mathbf{y}$ (the identity function) for HPCA, respectively. 
The second block is a \textit{reconstruction} function $\rho(\mathbf{y}, \mathbf{w})$, which aims at reconstructing the neurons' input, given their activations and weights. It performs the following computations: first, the activation feature map undergoes a transformation that depends on whether we are performing SWTA or HPCA; then, a matrix multiplication of the result by the transposed weight matrix is performed. Concerning the first step, focusing on neuron $j$, the transformation corresponds to setting the $j$-th channel to $1$ and the others to $0$, in the case of SWTA, or to the identity for the first $j$ channels and $0$ for the others, in the case of HPCA. This leads to $\rho(\mathbf{y}, \mathbf{w})_{i, j} = w_{i, j}$ for SWTA, and $\rho(\mathbf{y}, \mathbf{w})_{i, j} = \sum_{k=1}^{j} y_k w_{i, k}$ for HPCA, respectively (with indexes $i$ and $j$ running over all inputs and outputs). 
The third block is the Hebbian plasticity function, which computes a weight update, given the following three signals as inputs: (i) a \textit{target} signal (corresponding to the variable $x_i$), from which we want to discover patterns; (ii) a \textit{reconstruction} signal, derived from the internal representation of the neural layer through the reconstruction block; (iii) a \textit{gate}, derived from the neurons' activations through the gating block, that modulates the length of the weight update step. 

The weight update follows the direction of the difference between the target and reconstructed signals. 
Calling $\mathbf{s}$ the target signal, $\mathbf{s}*$ the reconstruction, and $\mathbf{g}$ the gate, a general Hebbian learning rule can be written as:
\begin{equation}\label{eq:hebb_general}
    \Delta w_{i, j} = \eta \, g_j \, ( s_i - s^*_{i, j} ) .
\end{equation}
In this newly introduced notation, $s_i$ corresponds to $x_i$, $s^*_i$ to $\rho(\mathbf{y}, \mathbf{w})_i$, and $g_j$ to $\gamma(\mathbf{y})_j$, respectively.

\paragraph{Patch-wise Hebbian learning and the shape mismatch problem} Let us instantiate Eq.~\ref{eq:hebb_general} in the specific case of standard convolutional layers. Let us call the feature map before the convolution the \textit{upsampled} feature map and the feature map after the convolution the \textit{downsampled} feature map. In the convolutional case, Hebbian updates are obtained by processing the input patch-wise, as shown in Fig.~\ref{fig:hebb_T-Conv_a_b}. In particular, the target signal corresponds to patches from the upsampled feature map, while reconstruction and gating terms are obtained from patches of the downsampled feature map through the respective blocks.

However, for T-Conv layers, we have a downsampled feature map before the layer and an upsampled feature map afterward. Simply applying the Hebbian learning rules in this scenario is not possible due to a \textit{shape mismatch} problem in the reconstruction block: the latter block transforms downsampled feature maps to upsampled feature maps, but this is not compatible with T-Conv layers, where we need instead to transform an upsampled feature map into a downsampled reconstruction. 
In order to solve this issue, we identified two possible strategies, which are shown in Fig.~\ref{fig:hebb_T-Conv_a_c} and Fig.~\ref{fig:hebb_T-Conv_a_d} and described in the following.

\begin{table}[!tb]
    \centering
    \caption{
    Summary of datasets. We report some statistics, an image sample, and, in the last row, the associated targets exploited during the supervised training step.
    }%
    \label{tab:datasets}%
    \renewcommand\tabularxcolumn[1]{m{#1}}%
    \newcolumntype{C}{>{\centering\arraybackslash}m{.23\linewidth}}%
    \setlength{\tabcolsep}{0pt}%
    \renewcommand*{\arraystretch}{0}%
    \newcommand{\imgwidth}{.23\linewidth}%
    \scriptsize%
    \begin{tabularx}{0.9\linewidth}{m{.17\linewidth}CCC}
        \toprule
        & GlaS~\cite{SIRINUKUNWATTANA2017489} & PH2~\cite{6610779} &        HMEPS~\cite{raffaele_mazziotti_2021_4488164} \\ [0.5ex]
        \midrule
        Images & 165 & 200 & 11,897 \\ [1ex]
        Size (px) & 775$\times$522 & 768$\times$560 & 128$\times$128 \\ [1ex]
        Segm. Task & Colorectal Cancer & Skin Lesion & Eye Pupil \\ 
        \midrule
        \multicolumn{1}{c}{\hspace{-0ex}\rotatebox[origin=c]{90}{Sample}} & \multicolumn{3}{c}{{%
            \setlength{\fboxsep}{0pt}%
            \setlength{\fboxrule}{1pt}%
            \fcolorbox{olive}{white}{\includegraphics[align=c,height=1.2cm,width=\imgwidth]{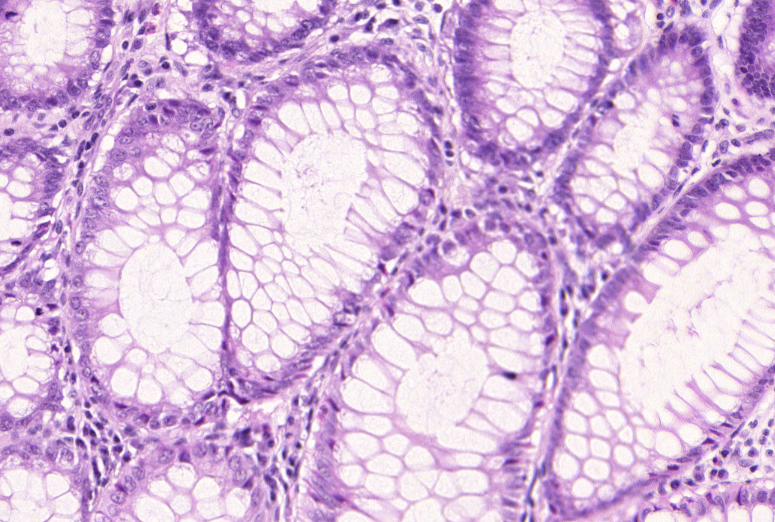}}%
            \fcolorbox{olive}{white}{\includegraphics[align=c,height=1.2cm,width=\imgwidth]{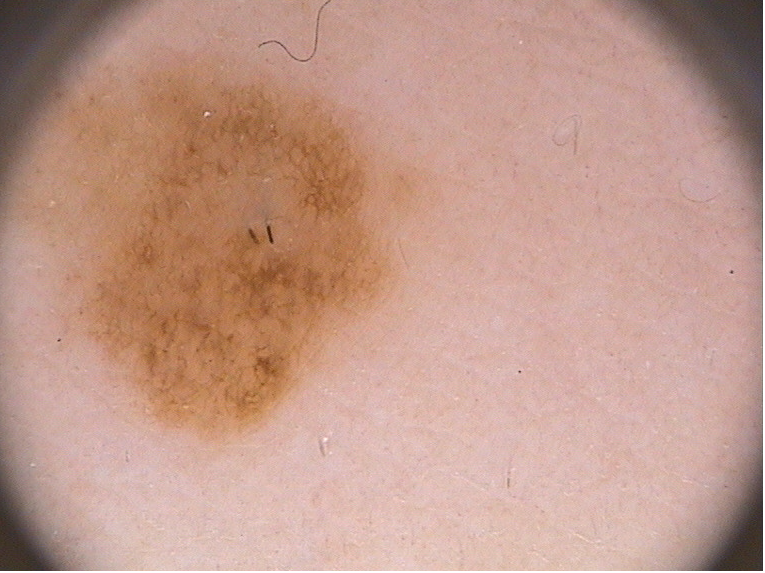}}%
            \fcolorbox{olive}{white}{\includegraphics[align=c,height=1.2cm,width=\imgwidth]{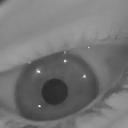}}%          
        }} \\ [6ex]
        %\midrule
        %
        \multicolumn{1}{c}{\hspace{-0ex}\rotatebox[origin=c]{90}{Target}} & \multicolumn{3}{c}{{%
            \setlength{\fboxsep}{0pt}%
            \setlength{\fboxrule}{1pt}%
            \fcolorbox{olive}{white}{\includegraphics[align=c,height=1.4cm,width=\imgwidth]{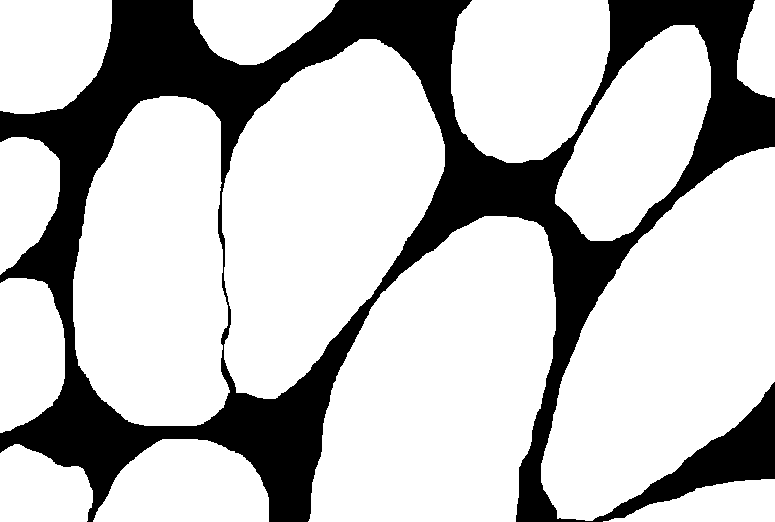}}%
            \fcolorbox{olive}{white}{\includegraphics[align=c,height=1.4cm,width=\imgwidth]{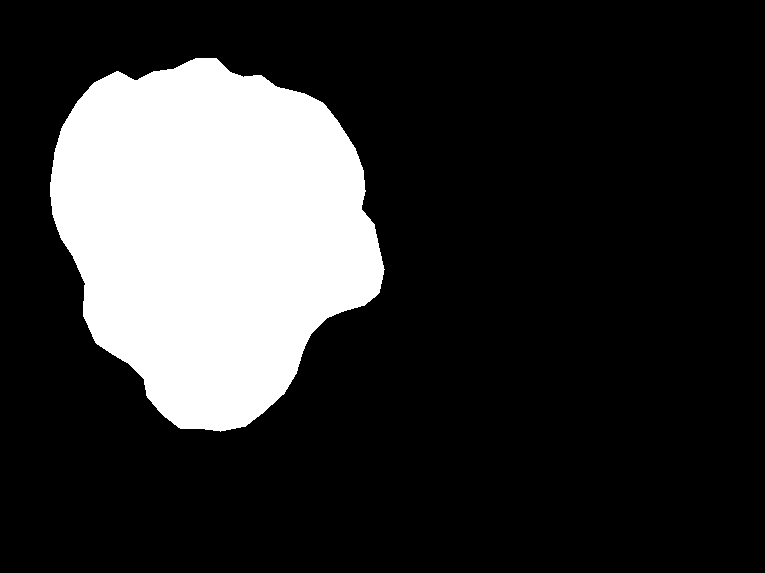}}%
            \fcolorbox{olive}{white}{\includegraphics[align=c,height=1.4cm,width=\imgwidth]{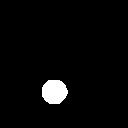}}%
        }} \\ 
        \bottomrule
    \end{tabularx}
\end{table}

\paragraph{First strategy: SWTA-S, HPCA-S} The first strategy is to exchange the role of downsampled and upsampled feature maps in the learning equations so that the same building blocks of Hebbian learning for standard convolutional layers can be reused in a different fashion. Specifically, as before, we use patches from the upsampled feature map as the target signal, while reconstruction and gating terms are computed from the corresponding patches in the downsampled feature map through the respective blocks.
This methodology for applying Hebbian learning in T-Conv layers is straightforward because it allows us to simply reuse the same building blocks in Fig.~\ref{fig:hebb_T-Conv_a} in these new settings. For this reason, we call the resulting learning methodologies following this \textit{Straightfarward} extension as SWTA-S and HPCA-S.

\paragraph{Second strategy: SWTA-TSA, HPCA-TSA} A drawback of the previous formulation is that the relationships with the unsupervised feature extraction principles that underlie the Hebbian pattern discovery processes (i.e.,, clustering for SWTA and PCA for HPCA) are lost. In fact, we are treating upsampled and downsampled feature maps as we would do for an ordinary convolutional block, although the relationship between inputs and outputs is reversed: the upsampled feature map is now the output of the neurons, and the downsampled feature map is the input.

Therefore, we also investigate a second strategy for applying Hebbian learning in T-Conv layers, where the reconstruction block is appropriately redesigned, in order to address the shape mismatch issue. 
In this case, as shown in Fig.~\ref{fig:hebb_T-Conv_a_d}, the downsampled feature map is treated as the target signal, while reconstruction and gating terms are computed from the upsampled feature map.
We introduce a new reconstruction block $\rho^*(\mathbf{y}, \mathbf{w})$ which performs the following computations: (i) apply a different transformation to the upsampled feature map, depending on whether we are performing HPCA or SWTA; (ii) extract patches at different offsets from the resulting feature maps, with the desired size, stride, etc. (also known as \textit{unfolding}); and (iii) perform matrix multiplication between the (vectorized) patches and the weight matrix.
The transformations mentioned in step (i) are as follows: concerning neuron $j$, in the case of SWTA, set the $j$-th channel to $1$ and the rest to $0$; in the case of HPCA, the transformation is the identity for the first $j$ channels and set the others to $0$.
Steps (ii) and (iii) are also equivalent to -- and can be implemented efficiently as -- an ordinary convolution. From now on, we call this \textit{Transposed-Structure-Aware} (TSA) formulation of SWTA and HPCA to T-Conv layers as SWTA-TSA and HPCA-TSA, respectively.

%In the following, our experimental comparison shows that this \textit{Transposed-Structure-Aware} (TSA) formulation is more effective than the straightforward formulation. From now on, we call the resulting extensions of SWTA and HPCA to T-Conv layers SWTA-TSA and HPCA-TSA.

%As already stated, we proceed in two stages.
%In the first stage, we employ SWTA or HPCA learning for updating the weights of the encoder, \ie, the standard convolutional layers of the downsampling path, as already done in~\cite{lagani2022c, lagani2024a}. Conversely, we use our novel SWTA-TSA or HPCA-TSA learning paradigms for updating the weights of the decoder, \ie, the T-Conv layers of the upsampling path. 
%In the second stage, we initialize the model with these weights, and we fine-tune the whole architecture with standard backpropagation. However, it is worth noting that this initialization can also be applied to other existing SOTA methods based on pseudo-labeling/consistency criteria.
%As shown in our experimental analysis, this leads to improved performance.  

In the experimental section, we focus on SWTA-based methods, which empirically lead to better results. Indeed, for semantic segmentation tasks where pixel-wise classification is required, clustering can be a good proxy for unsupervised class discovery in the observed tensors, therefore representing a promising inductive bias for successive fine-tuning. Specifically, we consider the setting involving SWTA-TSA, which resulted in the best-performing solution. However, we also report an ablation study over the different learning models (SWTA-S, HPCA-S, SWTA-TSA, HPCA-TSA).
%, and additional results can be found in Appendix~\ref{appendix:experiments}.

%%%%%%%%%%%%%%%%%%%%%%%%%%%%%%%%%%%%%%%%%%%%%%%%%%%%%%%%%%%%%%%%%%%%%%%%%%%%%%%
\section{Experimental Evaluation}
\label{sec:experiments}

%In this section, we describe the experimental settings and discuss the results. 
%comparing our approach based on SWTA-TSA with SOTA. Finally, we show additional experiments with existing semi-supervised methodologies initialized with our Hebbian unsupervised pre-training. 
%Qualitative results, implementation details, and additional analyses can be found in the Suppl. Material. 

\subsection{Datasets and Evaluation Metrics}
\label{sec:sec:datasets}
\noindent We assess our proposed approach on three public datasets widely used in the 2D biomedical image segmentation literature~\cite{9625988,10.1007/978-3-030-87193-2_4,10.1016/j.patcog.2022.108673,10332179} featuring different imaging modalities and segmentation tasks (see also Tab.~\ref{tab:datasets}).

\noindent \textbf{GlaS~\cite{SIRINUKUNWATTANA2017489}.} The Gland Segmentation in Colon Histology Images Challenge (GlaS) dataset includes 165 
%Hematoxylin and Eosin 
(H\&E) stained histological 775$\times$522 images of colorectal adenocarcinoma, of which 80 for training 
%(37 benign, 43 malignant) 
and 85 for testing.
%(37 benign, 48 malignant).

\noindent \textbf{PH2~\cite{6610779}.} This dermoscopic image database comprises 200 melanocytic lesions, including 80 common nevi, 80 atypical nevi, and 40 melanomas; they are 8-bit RGB color images with a resolution of 768$\times$560 pixels.
%, obtained through the Tuebinger Mole Analyzer system using a magnification of 20$\times$. 

%\noindent \textbf{LA~\cite{XIONG2021101832}.} The Left Atrial (LA) dataset contains 100 3D MRI images from the 2018 atrial segmentation challenge; following the literature, we use 80 images for training and 20 for testing.
\noindent \textbf{HMEPS~\cite{raffaele_mazziotti_2021_4488164}.} The Human and Mouse Eyes for Pupil Semantic Segmentation (HMEPS) dataset is a collection of 11,897 grayscale images of humans (4285) and mice (7612) eyes illuminated using infrared (IR, 850 nm) light sources, labeled 
%by five human annotators who manually placed a 
with polygons over pupil areas.

%In our results, 
We report four metrics widely used for biomedical image segmentation~\cite{LUO2022102517,Luo_2021,7785132,9706678}, i.e., Dice Coefficient (DC), Jaccard Index (JI), 95th percentile Hausdorff Distance (95HD), and Average Surface Distance (ASD). The first two evaluators emphasize pixel-wise accuracy, while 95HD and ASD focus on boundary accuracy. DC is considered the gold standard evaluator for this task.

\begin{table}[!tb]
    \centering
    \caption{Comparisons with SOTA on the GlaS dataset~\cite{SIRINUKUNWATTANA2017489}. \textcolor{Red}{Red} and \textcolor{Blue}{blue} indicate the best and second-best performance.
    %with 1\%, 2\%, 5\%, 10\%, and 20\% labeled data. 
    %Mean $\pm$ 90\% CI are reported. Best results are in red.
    }%
    \label{tab:comparison-sota-glas}%
    \newcolumntype{L}{>{\centering\arraybackslash}m{.1\linewidth}}%
    \newcolumntype{M}{>{\centering\arraybackslash}m{.17\linewidth}}%
    \newcolumntype{C}{>{\centering\arraybackslash}m{.13\linewidth}}
    \tiny%
    \setlength{\tabcolsep}{3pt}
    \begin{tabularx}{0.98\linewidth}{L|M|CCCC}
        \toprule
        Labeled \% & Method & DC (\%) $\uparrow$ & JI (\%) $\uparrow$ & 95HD $\downarrow$ & ASD $\downarrow$ \\
        \midrule
        100\% & Fully Sup. & 90.62 $\pm$ 0.20 & 82.85 $\pm$ 0.34 & 8.96 $\pm$ 0.34 & 1.76 $\pm$ 0.05 \\
        \midrule
        \midrule
        \multirow{8}{*}{1\%} & VAE~\cite{DBLP:journals/corr/KingmaW13} & 68.77 $\pm$ 0.51 & 52.42 $\pm$ 0.59 & 38.33 $\pm$ 5.68 & 11.90 $\pm$ 1.99 \\
        %& DDPM~\cite{10.5555/3495724.3496298} & - $\pm$ - & - $\pm$ - & - $\pm$ - & - $\pm$ - \\
        & SuperpixSSL~\cite{10.1007/978-3-030-58526-6_45} & 68.51 $\pm$ 0.50 & 52.11 $\pm$ 0.59 & 39.72 $\pm$ 5.59 & 12.48 $\pm$ 1.88 \\
        & EM~\cite{8954439} & 68.92 $\pm$ 0.77 & 52.60 $\pm$ 0.90 & 40.12 $\pm$ 3.48 & 10.92 $\pm$ 1.04 \\
        & CCT~\cite{9157032} & 68.97 $\pm$ 0.73 & 52.65 $\pm$ 0.86 & 40.72 $\pm$ 3.68 & 11.00 $\pm$ 1.21 \\
        %& MT~\cite{10.5555/3294771.3294885} & 64.63 $\pm$ 1.75 & 49.50 $\pm$ 1.88 & 56.24 $\pm$ 1.68 & 8.87 $\pm$ 0.45 \\
        & UAMT~\cite{10.1007/978-3-030-32245-8_67} & 69.12 $\pm$ 0.86 & 52.83 $\pm$ 1.02 & \textcolor{Blue}{\textbf{37.56 $\pm$ 4.76}} & \textcolor{Blue}{\textbf{10.24 $\pm$ 1.43}} \\
        %& LHFC~\cite{10376766} & 65.96 $\pm$ 0.22 & 51.10 $\pm$ 0.23 & 60.66 $\pm$ 1.41 & 9.36 $\pm$ 0.22 \\
        & CPS~\cite{9577639} & \textcolor{Blue}{\textbf{69.32 $\pm$ 0.59}} & \textcolor{Blue}{\textbf{53.05 $\pm$ 0.69}} & 38.29 $\pm$ 4.85 & 10.46 $\pm$ 1.37 \\
        & URPC~\cite{LUO2022102517} & 68.38 $\pm$ 0.44 & 51.96 $\pm$ 0.51 & 44.13 $\pm$ 2.92 & 12.04 $\pm$ 0.84 \\
        & \cellcolor{GreenYellow} \textbf{Ours} & \textcolor{Red}{\textbf{69.95 $\pm$ 1.09}} & \textcolor{Red}{\textbf{53.82 $\pm$ 1.31}} & \textcolor{Red}{\textbf{32.87 $\pm$ 3.36}} & \textcolor{Red}{\textbf{9.05 $\pm$ 1.09}} \\
        \midrule
        \multirow{8}{*}{2\%} & VAE~\cite{DBLP:journals/corr/KingmaW13} & 70.10 $\pm$ 0.95 & 53.99 $\pm$ 1.13 & 29.92 $\pm$ 4.05 & 8.88 $\pm$ 1.30 \\
        %& DDPM~\cite{10.5555/3495724.3496298} & - $\pm$ - & - $\pm$ - & - $\pm$ - & - $\pm$ - \\
        & SuperpixSSL~\cite{10.1007/978-3-030-58526-6_45} & 70.10 $\pm$ 1.25 & 54.00 $\pm$ 1.51 & 33.97 $\pm$ 5.02 & 10.24 $\pm$ 1.75 \\
        & EM~\cite{8954439} & 70.23 $\pm$ 1.34 & 54.16 $\pm$ 1.60 & 36.84 $\pm$ 4.92 & 9.65 $\pm$ 1.45 \\
        & CCT~\cite{9157032} & 70.05 $\pm$ 0.94 & 53.92 $\pm$ 1.11 & 31.26 $\pm$ 5.74 & 8.08 $\pm$ 1.71 \\
        %& MT~\cite{10.5555/3294771.3294885} & 66.46 $\pm$ 2.26 & 51.68 $\pm$ 2.54 & 48.90 $\pm$ 2.62 & 7.65 $\pm$ 0.43 \\
        & UAMT~\cite{10.1007/978-3-030-32245-8_67} & 69.71 $\pm$ 1.27 & 53.55 $\pm$ 1.52 & 35.35 $\pm$ 5.70 & 9.35 $\pm$ 1.72 \\
        %& LHFC~\cite{10376766} & 66.00 $\pm$ 0.07 & 51.14 $\pm$ 0.07 & 61.11 $\pm$ 0.60 & 9.41 $\pm$ 0.12 \\
        & CPS~\cite{9577639} & \textcolor{Blue}{\textbf{70.60 $\pm$ 1.13}} & \textcolor{Blue}{\textbf{54.59 $\pm$ 1.35}} & \textcolor{Red}{\textbf{29.81 $\pm$ 4.82}} & \textcolor{Red}{\textbf{7.57 $\pm$ 1.48}} \\
        & URPC~\cite{LUO2022102517} & 68.67 $\pm$ 0.98 & 52.31 $\pm$ 1.17 & 42.20 $\pm$ 3.21 & 11.51 $\pm$ 0.95 \\
        & \cellcolor{GreenYellow} \textbf{Ours} & \textcolor{Red}{\textbf{71.18 $\pm$ 1.28}} & \textcolor{Red}{\textbf{55.30 $\pm$ 1.54}} & \textcolor{Blue}{\textbf{30.07 $\pm$ 5.09}} & \textcolor{Blue}{\textbf{7.64 $\pm$ 1.56}} \\
        \midrule
        \multirow{8}{*}{5\%} & VAE~\cite{DBLP:journals/corr/KingmaW13} & 76.16 $\pm$ 1.21 & 61.54 $\pm$ 1.56 & 23.45 $\pm$ 3.15 & 6.33 $\pm$ 1.06 \\
        %& DDPM~\cite{10.5555/3495724.3496298} & - $\pm$ - & - $\pm$ - & - $\pm$ - & - $\pm$ - \\
        & SuperpixSSL~\cite{10.1007/978-3-030-58526-6_45} & 75.18 $\pm$ 1.27 & 60.28 $\pm$ 1.63 & 23.19 $\pm$ 2.55 & 6.21 $\pm$ 0.73 \\
        & EM~\cite{8954439} & 75.34 $\pm$ 0.92 & 60.45 $\pm$ 1.17 & 24.74 $\pm$ 2.21 & 5.86 $\pm$ 0.67 \\
        & CCT~\cite{9157032} & \textcolor{Blue}{\textbf{76.33 $\pm$ 1.10}} & \textcolor{Blue}{\textbf{61.76 $\pm$ 1.43}} & 23.62 $\pm$ 2.67 & 5.54 $\pm$ 0.72 \\
        %& MT~\cite{10.5555/3294771.3294885} & 70.31 $\pm$ 1.43 & 55.91 $\pm$ 1.70 & 50.33 $\pm$ 2.99 & 7.52 $\pm$ 0.56 \\
        & UAMT~\cite{10.1007/978-3-030-32245-8_67} & 75.14 $\pm$ 0.65 & 60.19 $\pm$ 0.83 & 25.98 $\pm$ 1.79 & 6.18 $\pm$ 0.50 \\
        %& LHFC~\cite{10376766} & 67.36 $\pm$ 2.38 & 52.81 $\pm$ 3.03 & 58.52 $\pm$ 3.59 & 9.07 $\pm$ 0.50 \\
        & CPS~\cite{9577639} & 76.17 $\pm$ 0.98 & 61.54 $\pm$ 1.28 & \textcolor{Blue}{\textbf{22.92 $\pm$ 1.84}} & \textcolor{Blue}{\textbf{5.28 $\pm$ 0.37}} \\
        & URPC~\cite{LUO2022102517} & 74.32 $\pm$ 1.06 & 59.17 $\pm$ 1.34 & 26.72 $\pm$ 2.79 & 6.59 $\pm$ 0.76 \\
        & \cellcolor{GreenYellow} \textbf{Ours} & \textcolor{Red}{\textbf{77.18 $\pm$ 1.05}} & \textcolor{Red}{\textbf{62.87 $\pm$ 1.40}} & \textcolor{Red}{\textbf{21.01 $\pm$ 2.47}} & \textcolor{Red}{\textbf{4.86 $\pm$ 0.60}} \\
        \midrule
        \multirow{8}{*}{10\%} & VAE~\cite{DBLP:journals/corr/KingmaW13} & 79.28 $\pm$ 1.39 & 65.73 $\pm$ 1.88 & 17.98 $\pm$ 1.35 & 4.66 $\pm$ 0.42 \\
        %& DDPM~\cite{10.5555/3495724.3496298} & - $\pm$ - & - $\pm$ - & - $\pm$ - & - $\pm$ - \\
        & SuperpixSSL~\cite{10.1007/978-3-030-58526-6_45} & 77.82 $\pm$ 1.67 & 63.78 $\pm$ 2.19 & 21.05 $\pm$ 3.43 & 5.67 $\pm$ 1.24 \\
        & EM~\cite{8954439} & 78.08 $\pm$ 0.82 & 64.06 $\pm$ 1.10 & 21.76 $\pm$ 1.81 & 4.87 $\pm$ 0.44 \\
        & CCT~\cite{9157032} & \textcolor{Blue}{\textbf{80.36 $\pm$ 1.05}} & \textcolor{Blue}{\textbf{67.19 $\pm$ 1.44}} & \textcolor{Blue}{\textbf{17.80 $\pm$ 1.05}} & \textcolor{Blue}{\textbf{3.91 $\pm$ 0.36}} \\
        %& MT~\cite{10.5555/3294771.3294885} & 74.49 $\pm$ 1.17 & 60.73 $\pm$ 1.54 & 43.92 $\pm$ 2.68 & 6.49 $\pm$ 0.42 \\
        & UAMT~\cite{10.1007/978-3-030-32245-8_67} & 79.31 $\pm$ 0.77 & 65.72 $\pm$ 1.04 & 18.37 $\pm$ 1.38 & 4.12 $\pm$ 0.25 \\
        %& LHFC~\cite{10376766} & 67.63 $\pm$ 1.89 & 53.05 $\pm$ 2.42 & 55.45 $\pm$ 4.62 & 8.61 $\pm$ 0.72 \\
        & CPS~\cite{9577639} & 80.35 $\pm$ 1.11 & 67.16 $\pm$ 1.56 & 18.73 $\pm$ 1.58 & 4.23 $\pm$ 0.46 \\
        & URPC~\cite{LUO2022102517} & 78.59 $\pm$ 1.39 & 64.78 $\pm$ 1.85 & 21.57 $\pm$ 3.05 & 5.03 $\pm$ 0.93 \\
        & \cellcolor{GreenYellow} \textbf{Ours} & \textcolor{Red}{\textbf{80.77 $\pm$ 0.77}} & \textcolor{Red}{\textbf{67.77 $\pm$ 1.08}} & \textcolor{Red}{\textbf{17.64 $\pm$ 0.82}} & \textcolor{Red}{\textbf{3.87 $\pm$ 0.21}} \\
        \midrule
        \multirow{8}{*}{20\%} & VAE~\cite{DBLP:journals/corr/KingmaW13} & 83.22 $\pm$ 1.06 & 71.30 $\pm$ 1.55 & 15.20 $\pm$ 0.82 & 3.61 $\pm$ 0.24 \\
        %& DDPM~\cite{10.5555/3495724.3496298} & - $\pm$ - & - $\pm$ - & - $\pm$ - & - $\pm$ - \\
        & SuperpixSSL~\cite{10.1007/978-3-030-58526-6_45} & 81.33 $\pm$ 1.36 & 68.60 $\pm$ 1.91 & 17.13 $\pm$ 1.54 & 4.16 $\pm$ 0.37 \\
        & EM~\cite{8954439} & 81.20 $\pm$ 0.80 & 68.38 $\pm$ 1.13 & 15.96 $\pm$ 1.11 & 3.55 $\pm$ 0.24 \\
        & CCT~\cite{9157032} & \textcolor{Blue}{\textbf{84.22 $\pm$ 0.84}} & \textcolor{Blue}{\textbf{72.76 $\pm$ 1.25}} & \textcolor{Blue}{\textbf{14.26 $\pm$ 0.79}} & \textcolor{Blue}{\textbf{2.98 $\pm$ 0.13}} \\
        %& MT~\cite{10.5555/3294771.3294885} & 78.14 $\pm$ 0.92 & 65.26 $\pm$ 1.11 & 40.01 $\pm$ 4.18 & 5.72 $\pm$ 0.76 \\
        & UAMT~\cite{10.1007/978-3-030-32245-8_67} & 83.03 $\pm$ 0.69 & 71.00 $\pm$ 1.00 & 14.56 $\pm$ 0.68 & 3.22 $\pm$ 0.19 \\
        %& LHFC~\cite{10376766} & 74.25 $\pm$ 3.38 & 61.41 $\pm$ 4.25 & 45.45 $\pm$ 11.01 & 7.14 $\pm$ 1.84 \\
        & CPS~\cite{9577639} & 83.90 $\pm$ 0.51 & 72.27 $\pm$ 0.77 & 14.29 $\pm$ 0.32 & 3.09 $\pm$ 0.11 \\
        & URPC~\cite{LUO2022102517} & 82.34 $\pm$ 2.07 & 70.12 $\pm$ 2.84 & 16.74 $\pm$ 2.16 & 3.64 $\pm$ 0.57 \\
        & \cellcolor{GreenYellow} \textbf{Ours} & \textcolor{Red}{\textbf{84.50 $\pm$ 0.50}} & \textcolor{Red}{\textbf{73.17 $\pm$ 0.75}} & \textcolor{Red}{\textbf{13.96 $\pm$ 0.55}} & \textcolor{Red}{\textbf{2.93 $\pm$ 0.11}} \\
        \bottomrule
    \end{tabularx}
\end{table}

\begin{table}[!tb]
    \centering
    \caption{Comparisons with SOTA on the PH2 dataset~\cite{6610779}. \textcolor{Red}{Red} and \textcolor{Blue}{blue} indicate the best and second-best performance.
    %with 1\%, 2\%, 5\%, 10\%, and 20\% labeled data. 
    %Mean $\pm$ 90\% CI are reported. Best results are in red.
    }%
    \label{tab:comparison-sota-ph2}%
    \newcolumntype{L}{>{\centering\arraybackslash}m{.1\linewidth}}%
    \newcolumntype{M}{>{\centering\arraybackslash}m{.17\linewidth}}%
    \newcolumntype{C}{>{\centering\arraybackslash}m{.13\linewidth}}
    \tiny%
    \setlength{\tabcolsep}{3pt}
    \begin{tabularx}{0.98\linewidth}{L|M|CCCC}
        \toprule
        Labeled \% & Method & DC (\%) $\uparrow$ & JI (\%) $\uparrow$ & 95HD $\downarrow$ & ASD $\downarrow$ \\
        \midrule
        %\midrule
        100\% & Fully Sup. & 92.44 $\pm$ 0.38 & 85.96 $\pm$ 0.66 & 6.77 $\pm$ 0.72 & 2.34 $\pm$ 0.12 \\
        \midrule
        \midrule
        %\multirow{6}{*}{1\%} & UNet~\cite{10.1007/978-3-319-24574-4_28} & 73.76 $\pm$ 2.92 & \textcolor{Red}{\underline{\textbf{62.45 $\pm$ 3.43}}} & 49.30 $\pm$ 13.63 & 8.76 $\pm$ 2.68 \\
        \multirow{8}{*}{1\%} & VAE~\cite{DBLP:journals/corr/KingmaW13} & 75.83 $\pm$ 0.96 & 61.09 $\pm$ 1.22 & 32.37 $\pm$ 6.10 & 10.44 $\pm$ 2.05 \\
        %& DDPM~\cite{10.5555/3495724.3496298} & - $\pm$ - & - $\pm$ - & - $\pm$ - & - $\pm$ - \\
        & SuperpixSSL~\cite{10.1007/978-3-030-58526-6_45} & 70.78 $\pm$ 2.12 & 54.87 $\pm$ 2.60 & 33.79 $\pm$ 9.46 & 12.66 $\pm$ 3.01 \\
        & EM~\cite{8954439} & 73.24 $\pm$ 2.32 & 57.92 $\pm$ 2.87 & 25.96 $\pm$ 7.74 & 10.01 $\pm$ 3.03 \\
        & CCT~\cite{9157032} & 73.42 $\pm$ 1.58 & 58.06 $\pm$ 1.98 & 27.40 $\pm$ 7.09 & 10.27 $\pm$ 2.53 \\
        %& MT~\cite{10.5555/3294771.3294885} & 68.75 $\pm$ 4.81 & 57.09 $\pm$ 5.20 & 83.46 $\pm$ 19.53 & 14.83 $\pm$ 3.22 \\
        %& LHFC~\cite{10376766} & 62.54 $\pm$ 2.48 & 49.10 $\pm$ 2.85 & 119.69 $\pm$ 4.26 & 20.86 $\pm$ 1.22 \\
        & UAMT~\cite{10.1007/978-3-030-32245-8_67} & 74.72 $\pm$ 1.45 & 59.70 $\pm$ 1.84 & 25.06 $\pm$ 6.78 & \textcolor{Blue}{\textbf{8.20 $\pm$ 1.31}} \\
        & CPS~\cite{9577639} & \textcolor{Blue}{\textbf{76.07 $\pm$ 2.12}} & \textcolor{Blue}{\textbf{61.50 $\pm$ 2.71}} & 28.79 $\pm$ 7.93 & 10.03 $\pm$ 2.81 \\
        & URPC~\cite{LUO2022102517} & 71.23 $\pm$ 1.95 & 55.39 $\pm$ 2.33 & \textcolor{Red}{\textbf{23.17 $\pm$ 8.10}} & 10.41 $\pm$ 2.59 \\
        & \cellcolor{GreenYellow} \textbf{Ours} & \textcolor{Red}{\textbf{78.16 $\pm$ 3.04}} & \textcolor{Red}{\textbf{64.25 $\pm$ 3.69}} & \textcolor{Blue}{\textbf{24.19 $\pm$ 8.57}} & \textcolor{Red}{\textbf{8.18 $\pm$ 2.39}} \\
        \midrule
        %\multirow{6}{*}{2\%} & UNet~\cite{10.1007/978-3-319-24574-4_28} & 74.48 $\pm$ 3.56 & 62.33 $\pm$ 4.24 & 36.84 $\pm$ 10.25 & 5.92 $\pm$ 1.71 \\
        \multirow{8}{*}{2\%} & VAE~\cite{DBLP:journals/corr/KingmaW13} & 78.78 $\pm$ 1.61 & 65.06 $\pm$ 2.18 & 22.29 $\pm$ 4.73 & 7.24 $\pm$ 1.06 \\
        %& DDPM~\cite{10.5555/3495724.3496298} & - $\pm$ - & - $\pm$ - & - $\pm$ - & - $\pm$ - \\
        & SuperpixSSL~\cite{10.1007/978-3-030-58526-6_45} & 77.61 $\pm$ 2.47 & 63.59 $\pm$ 3.28 & 19.13 $\pm$ 1.88 & 7.16 $\pm$ 0.72 \\
        & EM~\cite{8954439} & 79.34 $\pm$ 2.63 & 65.95 $\pm$ 3.55 & 17.69 $\pm$ 3.41 & 6.62 $\pm$ 0.96 \\
        & CCT~\cite{9157032} & \textcolor{Blue}{\textbf{80.13 $\pm$ 1.41}} & \textcolor{Blue}{\textbf{66.90 $\pm$ 1.96}} & \textcolor{Red}{\textbf{13.73 $\pm$ 0.88}} & \textcolor{Red}{\textbf{5.82 $\pm$ 0.47}} \\
        %& MT~\cite{10.5555/3294771.3294885} & 75.94 $\pm$ 3.48 & 64.77 $\pm$ 4.02 & 70.76 $\pm$ 17.54 & 10.95 $\pm$ 3.02 \\
        %& LHFC~\cite{10376766} & 71.11 $\pm$ 4.30 & 57.59 $\pm$ 5.84 & 98.17 $\pm$ 62.67 & 16.57 $\pm$ 11.41 \\
        & UAMT~\cite{10.1007/978-3-030-32245-8_67} & 79.76 $\pm$ 1.26 & 66.38 $\pm$ 1.74 & 15.05 $\pm$ 1.89 & 6.10 $\pm$ 0.59 \\
        & CPS~\cite{9577639} & 79.64 $\pm$ 2.72 & 66.39 $\pm$ 3.71 & 14.97 $\pm$ 1.92 & 6.33 $\pm$ 0.88 \\
        & URPC~\cite{LUO2022102517} & 78.88 $\pm$ 1.69 & 65.22 $\pm$ 2.33 & \textcolor{Blue}{\textbf{14.68 $\pm$ 1.33}} & 6.90 $\pm$ 0.76 \\
        & \cellcolor{GreenYellow} \textbf{Ours} & \textcolor{Red}{\textbf{80.90 $\pm$ 1.66}} & \textcolor{Red}{\textbf{67.97 $\pm$ 2.36}} & 17.33 $\pm$ 2.30 & \textcolor{Blue}{\textbf{6.08 $\pm$ 0.38}} \\
        \midrule
        %\multirow{6}{*}{5\%} & UNet~\cite{10.1007/978-3-319-24574-4_28} & 77.06 $\pm$ 2.17 & 65.02 $\pm$ 2.81 & 30.05 $\pm$ 7.13  & 4.67 $\pm$ 1.20 \\
        \multirow{8}{*}{5\%} & VAE~\cite{DBLP:journals/corr/KingmaW13} & 82.00 $\pm$ 0.94 & 69.53 $\pm$ 1.33 & 18.33 $\pm$ 2.23 & 6.02 $\pm$ 0.54 \\
        %& DDPM~\cite{10.5555/3495724.3496298} & - $\pm$ - & - $\pm$ - & - $\pm$ - & - $\pm$ - \\
        & SuperpixSSL~\cite{10.1007/978-3-030-58526-6_45} & 81.78 $\pm$ 1.38 & 69.23 $\pm$ 1.95 & 15.41 $\pm$ 2.18 & 5.74 $\pm$ 0.57 \\
        & EM~\cite{8954439} & 81.89 $\pm$ 1.02 & 69.37 $\pm$ 1.44 & 13.72 $\pm$ 1.85 & 5.45 $\pm$ 0.34 \\
        & CCT~\cite{9157032} & 82.78 $\pm$ 1.15 & 70.66 $\pm$ 1.63 & 13.64 $\pm$ 1.71 & 5.27 $\pm$ 0.35 \\
        %& MT~\cite{10.5555/3294771.3294885} & 81.01 $\pm$ 2.81 & 70.81 $\pm$ 3.53 & 37.17 $\pm$ 12.66 & 5.67 $\pm$ 1.89 \\
        %& LHFC~\cite{10376766} & 78.62 $\pm$ 4.80 & 66.95 $\pm$ 6.00 & 35.38 $\pm$ 4.44 & 5.40 $\pm$ 0.89 \\
        & UAMT~\cite{10.1007/978-3-030-32245-8_67} & \textcolor{Blue}{\textbf{83.75 $\pm$ 1.11}} & \textcolor{Blue}{\textbf{72.08 $\pm$ 1.64}} & \textcolor{Blue}{\textbf{12.29 $\pm$ 0.77}} & \textcolor{Blue}{\textbf{4.81 $\pm$ 0.26}} \\
        & CPS~\cite{9577639} & 82.86 $\pm$ 1.18 & 70.78 $\pm$ 1.73 & 15.00 $\pm$ 3.41 & 5.53 $\pm$ 0.68 \\
        & URPC~\cite{LUO2022102517} & 83.41 $\pm$ 2.17 & 71.65 $\pm$ 3.14 & \textcolor{Red}{\textbf{10.78 $\pm$ 0.99}} & 4.86 $\pm$ 0.43 \\
        & \cellcolor{GreenYellow} \textbf{Ours} & \textcolor{Red}{\textbf{85.77 $\pm$ 1.51}} & \textcolor{Red}{\textbf{75.13 $\pm$ 2.28}} & 14.10 $\pm$ 0.87 & \textcolor{Red}{\textbf{4.78 $\pm$ 0.30}} \\
        \midrule
        %\multirow{6}{*}{10\%} & UNet~\cite{10.1007/978-3-319-24574-4_28} & 80.05 $\pm$ 2.13 & 68.52 $\pm$ 2.56 & 19.37 $\pm$ 6.96 & 3.02 $\pm$ 1.17 \\
        \multirow{8}{*}{10\%} & VAE~\cite{DBLP:journals/corr/KingmaW13} & 84.29 $\pm$ 1.44 & 72.91 $\pm$ 2.15 & 15.06 $\pm$ 2.18 & 5.22 $\pm$ 0.61 \\
        %& DDPM~\cite{10.5555/3495724.3496298} & - $\pm$ - & - $\pm$ - & - $\pm$ - & - $\pm$ - \\
        & SuperpixSSL~\cite{10.1007/978-3-030-58526-6_45} & 85.21 $\pm$ 1.15 & 74.28 $\pm$ 1.77 & 12.27 $\pm$ 1.42 & 4.53 $\pm$ 0.28 \\
        & EM~\cite{8954439} & 84.94 $\pm$ 0.90 & 73.85 $\pm$ 1.35 & \textcolor{Blue}{\textbf{11.21 $\pm$ 1.41}} & 4.48 $\pm$ 0.41 \\
        & CCT~\cite{9157032} & 87.93 $\pm$ 1.96 & 78.46 $\pm$ 3.14 & 13.05 $\pm$ 3.50 & 3.88 $\pm$ 1.49 \\
        %& MT~\cite{10.5555/3294771.3294885} & 84.61 $\pm$ 1.68 & 74.87 $\pm$ 2.35 & 23.56 $\pm$ 3.70 & 3.21 $\pm$ 0.47 \\
        %& LHFC~\cite{10376766} & 81.49 $\pm$ 1.76 & 70.62 $\pm$ 2.04 & 22.43 $\pm$ 4.30 & 3.24 $\pm$ 0.83 \\
        & UAMT~\cite{10.1007/978-3-030-32245-8_67} & 87.37 $\pm$ 0.58 & 77.59 $\pm$ 0.92 & 12.17 $\pm$ 1.64 & 4.43 $\pm$ 0.25 \\
        & CPS~\cite{9577639} & 84.33 $\pm$ 0.82 & 72.93 $\pm$ 1.21 & 13.19 $\pm$ 1.71 & 4.85 $\pm$ 0.37 \\
        & URPC~\cite{LUO2022102517} & \textcolor{Blue}{\textbf{88.06 $\pm$ 0.40}} & \textcolor{Blue}{\textbf{78.89 $\pm$ 0.45}} & \textcolor{Red}{\textbf{8.95 $\pm$ 1.19}} & \textcolor{Red}{\textbf{3.71 $\pm$ 0.36}} \\
        & \cellcolor{GreenYellow} \textbf{Ours} & \textcolor{Red}{\textbf{88.26 $\pm$ 0.51}} & \textcolor{Red}{\textbf{79.00 $\pm$ 0.82}} & 13.04 $\pm$ 2.39 & \textcolor{Blue}{\textbf{4.35 $\pm$ 0.63}} \\
        \midrule
        %\multirow{6}{*}{20\%} & UNet~\cite{10.1007/978-3-319-24574-4_28} & 84.17 $\pm$ 1.27 & 73.53 $\pm$ 1.77 & 11.11 $\pm$ 3.23 & 1.44 $\pm$ 0.30 \\
        \multirow{8}{*}{20\%} & VAE~\cite{DBLP:journals/corr/KingmaW13} & 87.93 $\pm$ 1.19 & 78.52 $\pm$ 1.90 & 12.13 $\pm$ 2.11 & 4.12 $\pm$ 0.58 \\
        %& DDPM~\cite{10.5555/3495724.3496298} & - $\pm$ - & - $\pm$ - & - $\pm$ - & - $\pm$ - \\
        & SuperpixSSL~\cite{10.1007/978-3-030-58526-6_45} & 88.31 $\pm$ 0.94 & 79.10 $\pm$ 1.52 & 10.39 $\pm$ 1.05 & 3.75 $\pm$ 0.27 \\
        & EM~\cite{8954439} & 86.30 $\pm$ 0.87 & 75.92 $\pm$ 1.33 & 10.33 $\pm$ 1.36 & 3.97 $\pm$ 0.33 \\
        & CCT~\cite{9157032} & 89.95 $\pm$ 0.57 & 81.74 $\pm$ 0.94 & 8.21 $\pm$ 0.60 & 3.04 $\pm$ 0.10 \\
        %& MT~\cite{10.5555/3294771.3294885} & 83.55 $\pm$ 1.18 & 73.35 $\pm$ 1.66 & 27.03 $\pm$ 4.67 & 3.64 $\pm$ 0.62 \\
        %& LHFC~\cite{10376766} & 84.96 $\pm$ 0.53 & 74.82 $\pm$ 0.85 & 17.03 $\pm$ 2.65 & 2.17 $\pm$ 0.32 \\
        & UAMT~\cite{10.1007/978-3-030-32245-8_67} & 88.95 $\pm$ 0.64 & 80.12 $\pm$ 1.04 & 10.52 $\pm$ 1.58 & 3.56 $\pm$ 0.28 \\
        & CPS~\cite{9577639} & 86.49 $\pm$ 0.97 & 76.23 $\pm$ 1.52 & 10.56 $\pm$ 1.15 & 4.14 $\pm$ 0.37 \\
        & URPC~\cite{LUO2022102517} & \textcolor{Blue}{\textbf{91.73 $\pm$ 0.80}} & \textcolor{Blue}{\textbf{84.71 $\pm$ 1.36}} & \textcolor{Red}{\textbf{6.35 $\pm$ 0.13}} & \textcolor{Blue}{\textbf{2.48 $\pm$ 0.11}} \\
        & \cellcolor{GreenYellow} \textbf{Ours} & \textcolor{Red}{\textbf{91.99 $\pm$ 0.82}} & \textcolor{Red}{\textbf{84.92 $\pm$ 1.38}} & \textcolor{Blue}{\textbf{8.01 $\pm$ 2.13}} & \textcolor{Red}{\textbf{2.13 $\pm$ 0.87}} \\
        \bottomrule
    \end{tabularx}
\end{table}

\subsection{Comparison with SOTA}
\label{sec:sec:results-comparison-sota}
\noindent We quantitatively compare our approach against several SOTA single-stage semi-supervised techniques comprising pseudo-labeling and consistency training strategies -- EM~\cite{8954439}, CCT~\cite{9157032}, UAMT~\cite{10.1007/978-3-030-32245-8_67}, CPS~\cite{9577639}, and URPC~\cite{LUO2022102517}. %Specifically, we face up against two pseudo-labeling approaches, \ie, EM~\cite{8954439},  %MT~\cite{10.5555/3294771.3294885}
%UAMT~\cite{10.1007/978-3-030-32245-8_67}, and two consistency training strategies, \ie, CCT~\cite{9157032} and 
%LHFC~\cite{10376766}. 
Since these SOTA methodologies have different experimental settings, we reimplement them to run the experimental evaluation under fair conditions, using the same UNet model as the underlying architecture. Moreover, we designed some two-stage pipelines as additional competitors, including unsupervised stages leveraging VAEs~\cite{DBLP:journals/corr/KingmaW13}, %DDPMs~\cite{10.5555/3495724.3496298}, 
and SSL~\cite{10.1007/978-3-030-58526-6_45}, all of them followed by supervised fine-tuning with backpropagation~\cite{kingma2014b}.
We performed experiments considering several semi-supervised setups, i.e., we supervised the models with 1\%, 2\%, 5\%, 10\%, and 20\% labeled images. We report results showing the mean over ten independent runs together with 90\% confidence intervals. Qualitative results can be found in Fig.~\ref{fig:qualitatite-results}.

\paragraph{GlaS} The results on the GlaS dataset are given in Tab.~\ref{tab:comparison-sota-glas}. Our approach outperforms previous works considering all the metrics almost in all the considered settings. Concerning DC, we outperform SOTA techniques by about 1\%-2\%, depending on the considered data regime.

\paragraph{PH2} Tab.~\ref{tab:comparison-sota-ph2} illustrates the results on the PH2 dataset. Even in this case, our approach achieves the best results in almost all the settings, except for 95HD and ASD in some training data regimes where, anyway, we obtain the second highest results. Concerning DC, we outperform SOTA techniques up to 3\%, depending on the considered data regime. 

\begin{table}[!tb]
    \centering
    \caption{Comparisons with SOTA on the HMEPS dataset~\cite{raffaele_mazziotti_2021_4488164}. \textcolor{Red}{Red} and \textcolor{Blue}{blue} indicate the best and second-best performance. 
    %with 1\%, 2\%, 5\%, 10\%, and 20\% labeled data. 
    %Mean $\pm$ 90\% CI are reported. Best results are in red.
    }%
    \label{tab:comparison-sota-hmeps}%
    \newcolumntype{L}{>{\centering\arraybackslash}m{.1\linewidth}}%
    \newcolumntype{M}{>{\centering\arraybackslash}m{.17\linewidth}}%
    \newcolumntype{C}{>{\centering\arraybackslash}m{.13\linewidth}}
    \tiny%
    \setlength{\tabcolsep}{3pt}
    \begin{tabularx}{0.98\linewidth}{L|M|CCCC}
        \toprule
        Labeled \% & Method & DC (\%) $\uparrow$ & JI (\%) $\uparrow$ & 95HD $\downarrow$ & ASD $\downarrow$ \\
        \midrule
        100\% & Fully Sup. & 96.98 $\pm$ 0.42 & 94.70 $\pm$ 0.43 & 0.06 $\pm$ 0.05 & 0.03 $\pm$ 0.00 \\
        \midrule
        \midrule
        \multirow{8}{*}{1\%} & VAE~\cite{DBLP:journals/corr/KingmaW13} & 89.53 $\pm$ 1.94 & 82.39 $\pm$ 3.30 & 7.14 $\pm$ 5.69 & 2.07 $\pm$ 2.36 \\
        %& DDPM~\cite{10.5555/3495724.3496298} & - $\pm$ - & - $\pm$ - & - $\pm$ - & - $\pm$ - \\
        & SuperpixSSL~\cite{10.1007/978-3-030-58526-6_45} & 87.45 $\pm$ 3.13 & 77.80 $\pm$ 4.94 & 18.67 $\pm$ 9.46 & 3.91 $\pm$ 3.20 \\
        & EM~\cite{8954439} & 90.24 $\pm$ 2.74 & 82.25 $\pm$ 4.52 & 3.95 $\pm$ 2.29 & 1.24 $\pm$ 0.72 \\
        & CCT~\cite{9157032} & \textcolor{Red}{\textbf{91.09 $\pm$ 2.50}} & \textcolor{Red}{\textbf{84.03 $\pm$ 4.24}} & \textcolor{Red}{\textbf{2.63 $\pm$ 0.51}} & \textcolor{Red}{\textbf{0.85 $\pm$ 0.46}} \\
        & UAMT~\cite{10.1007/978-3-030-32245-8_67} & 90.18 $\pm$ 0.77 & 82.12 $\pm$ 1.27 & 4.19 $\pm$ 2.20 & 1.15 $\pm$ 0.39 \\
        & CPS~\cite{9577639} & 90.39 $\pm$ 0.55 & 82.49 $\pm$ 0.91 & 4.40 $\pm$ 0.78 & 1.16 $\pm$ 0.11 \\
        & URPC~\cite{LUO2022102517} & 89.15 $\pm$ 0.79 & 80.45 $\pm$ 1.28 & 5.96 $\pm$ 1.92 & 1.46 $\pm$ 0.32 \\
        %& MT~\cite{10.5555/3294771.3294885} & 85.73 $\pm$ 1.27 & 78.83 $\pm$ 1.68 & 5.79 $\pm$ 2.33 & 0.75 $\pm$ 0.36 \\
        %& LHFC~\cite{10376766} & 92.22 $\pm$ 1.01 & 88.10 $\pm$ 1.15 & \textcolor{Red}{\underline{\textbf{0.45 $\pm$ 0.09}}} & \textcolor{Red}{\underline{\textbf{0.07 $\pm$ 0.00}}} \\
        & \cellcolor{GreenYellow} \textbf{Ours} & \textcolor{Blue}{\textbf{90.75 $\pm$ 0.50}} & \textcolor{Blue}{\textbf{83.07 $\pm$ 2.19}} & \textcolor{Blue}{\textbf{3.70 $\pm$ 1.45}} & \textcolor{Blue}{\textbf{1.07 $\pm$ 0.51}} \\
        \midrule
        \multirow{8}{*}{2\%} & VAE~\cite{DBLP:journals/corr/KingmaW13} & 91.51 $\pm$ 2.42 & 84.59 $\pm$ 3.77 & 3.82 $\pm$ 2.44 & 1.10 $\pm$ 0.64 \\
        %& DDPM~\cite{10.5555/3495724.3496298} & - $\pm$ - & - $\pm$ - & - $\pm$ - & - $\pm$ - \\
        & SuperpixSSL~\cite{10.1007/978-3-030-58526-6_45} & 88.47 $\pm$ 4.53 & 79.53 $\pm$ 7.16 & 21.04 $\pm$ 10.75 & 5.26 $\pm$ 3.34 \\
        & EM~\cite{8954439} & 91.35 $\pm$ 1.93 & 84.14 $\pm$ 3.19 & 2.72 $\pm$ 0.61 & 0.82 $\pm$ 0.23 \\
        & CCT~\cite{9157032} & 91.48 $\pm$ 2.45 & 84.34 $\pm$ 4.19 & 2.48 $\pm$ 1.53 & 0.83 $\pm$ 0.41 \\
        %& MT~\cite{10.5555/3294771.3294885} & 88.58 $\pm$ 1.79 & 82.69 $\pm$ 2.37 & 3.21 $\pm$ 1.78 & 0.45 $\pm$ 0.22 \\
        %& LHFC~\cite{10376766} & 93.50 $\pm$ 0.99 & 89.74 $\pm$ 1.20 & \textcolor{Red}{\underline{\textbf{0.10 $\pm$ 0.06}}} & \textcolor{Red}{\underline{\textbf{0.04 $\pm$ 0.00}}} \\
        & UAMT~\cite{10.1007/978-3-030-32245-8_67} & \textcolor{Blue}{\textbf{92.42 $\pm$ 0.09}} & \textcolor{Blue}{\textbf{86.06 $\pm$ 0.16}} & \textcolor{Blue}{\textbf{2.35 $\pm$ 0.90}} & \textcolor{Blue}{\textbf{0.72 $\pm$ 0.13}} \\
        & CPS~\cite{9577639} & 91.07 $\pm$ 0.28 & 83.60 $\pm$ 0.48 & 3.11 $\pm$ 0.81 & 0.93 $\pm$ 0.11 \\
        & URPC~\cite{LUO2022102517} & 89.78 $\pm$ 0.64 & 81.47 $\pm$ 1.05 & 3.99 $\pm$ 0.63 & 1.14 $\pm$ 0.11 \\
        & \cellcolor{GreenYellow} \textbf{Ours} & \textcolor{Red}{\textbf{92.60 $\pm$ 1.20}} & \textcolor{Red}{\textbf{86.21 $\pm$ 2.09}} & \textcolor{Red}{\textbf{2.25 $\pm$ 0.69}} & \textcolor{Red}{\textbf{0.70 $\pm$ 0.11}} \\
        \midrule
        \multirow{8}{*}{5\%} & VAE~\cite{DBLP:journals/corr/KingmaW13} & 93.09 $\pm$ 0.56 & 87.09 $\pm$ 0.98 & 2.10 $\pm$ 0.27 & 0.65 $\pm$ 0.07 \\
        %& DDPM~\cite{10.5555/3495724.3496298} & - $\pm$ - & - $\pm$ - & - $\pm$ - & - $\pm$ - \\
        & SuperpixSSL~\cite{10.1007/978-3-030-58526-6_45} & 91.34 $\pm$ 2.35 & 84.16 $\pm$ 3.96 & 4.36 $\pm$ 3.10 & 1.25 $\pm$ 0.71 \\
        & EM~\cite{8954439} & 92.31 $\pm$ 1.07 & 85.78 $\pm$ 1.84 & 2.38 $\pm$ 0.72 & 0.79 $\pm$ 0.17 \\
        & CCT~\cite{9157032} & \textcolor{Blue}{\textbf{93.25 $\pm$ 0.92}} & \textcolor{Blue}{\textbf{87.38 $\pm$ 1.57}} & \textcolor{Blue}{\textbf{1.59 $\pm$ 0.36}} & \textcolor{Blue}{\textbf{0.59 $\pm$ 0.11}} \\
        %& MT~\cite{10.5555/3294771.3294885} & 90.37 $\pm$ 1.32 & 84.70 $\pm$ 1.79 & 1.60 $\pm$ 0.60 & 0.89 $\pm$ 0.32 \\
        %& LHFC~\cite{10376766} & 94.82 $\pm$ 0.18 & 92.07 $\pm$ 0.44 & 0.09 $\pm$ 0.06 & 0.04 $\pm$ 0.00 \\
        & UAMT~\cite{10.1007/978-3-030-32245-8_67} & 93.03 $\pm$ 1.18 & 87.02 $\pm$ 2.00 & 2.08 $\pm$ 0.93 & 0.65 $\pm$ 0.16 \\
        & CPS~\cite{9577639} & 92.89 $\pm$ 0.41 & 86.72 $\pm$ 0.72 & 2.25 $\pm$ 0.79 & 0.69 $\pm$ 0.14 \\
        & URPC~\cite{LUO2022102517} & 90.85 $\pm$ 0.72 & 83.23 $\pm$ 1.22 & 4.11 $\pm$ 2.84 & 1.05 $\pm$ 0.39 \\
        & \cellcolor{GreenYellow} \textbf{Ours} & \textcolor{Red}{\textbf{93.51 $\pm$ 0.25}} & \textcolor{Red}{\textbf{87.81 $\pm$ 0.45}} & \textcolor{Red}{\textbf{1.35 $\pm$ 0.31}} & \textcolor{Red}{\textbf{0.47 $\pm$ 0.03}} \\
        \midrule
        \multirow{8}{*}{10\%} & VAE~\cite{DBLP:journals/corr/KingmaW13} & 93.38 $\pm$ 0.37 & 87.58 $\pm$ 0.65 & 1.72 $\pm$ 0.34 & 0.59 $\pm$ 0.09 \\
        %& DDPM~\cite{10.5555/3495724.3496298} & - $\pm$ - & - $\pm$ - & - $\pm$ - & - $\pm$ - \\
        & SuperpixSSL~\cite{10.1007/978-3-030-58526-6_45} & 92.82 $\pm$ 2.01 & 86.65 $\pm$ 3.44 & 2.94 $\pm$ 2.51 & 0.84 $\pm$ 0.62 \\
        & EM~\cite{8954439} & 92.56 $\pm$ 0.96 & 86.20 $\pm$ 1.65 & 2.42 $\pm$ 0.85 & 0.75 $\pm$ 0.19 \\
        & CCT~\cite{9157032} & \textcolor{Blue}{\textbf{93.45 $\pm$ 0.05}} & \textcolor{Blue}{\textbf{87.70 $\pm$ 0.09}} & \textcolor{Blue}{\textbf{1.55 $\pm$ 0.79}} & \textcolor{Blue}{\textbf{0.57 $\pm$ 0.18}} \\
        & UAMT~\cite{10.1007/978-3-030-32245-8_67} & 93.14 $\pm$ 0.47 & 87.16 $\pm$ 0.82 & 1.67 $\pm$ 0.43 & 0.58 $\pm$ 0.08 \\
        & CPS~\cite{9577639} & 92.83 $\pm$ 0.46 & 86.63 $\pm$ 0.81 & 1.83 $\pm$ 0.37 & 0.62 $\pm$ 0.08 \\
        & URPC~\cite{LUO2022102517} & 90.96 $\pm$ 0.83 & 83.44 $\pm$ 1.40 & 2.31 $\pm$ 0.58 & 0.83 $\pm$ 0.13 \\
        %& MT~\cite{10.5555/3294771.3294885} & 92.66 $\pm$ 1.03 & 87.93 $\pm$ 1.63 & 1.80 $\pm$ 1.25 & 0.23 $\pm$ 0.14 \\
        %& LHFC~\cite{10376766} & 95.52 $\pm$ 0.26 & 92.97 $\pm$ 0.34 & 0.08 $\pm$ 0.03 & 0.04 $\pm$ 0.00 \\
        & \cellcolor{GreenYellow} \textbf{Ours} & \textcolor{Red}{\textbf{93.68 $\pm$ 0.28}} & \textcolor{Red}{\textbf{88.12 $\pm$ 0.50}} & \textcolor{Red}{\textbf{1.37 $\pm$ 0.19}} & \textcolor{Red}{\textbf{0.49 $\pm$ 0.04}} \\
        \midrule
        \multirow{8}{*}{20\%} & VAE~\cite{DBLP:journals/corr/KingmaW13} & 93.42 $\pm$ 0.19 & 87.66 $\pm$ 0.34 & 1.78 $\pm$ 0.23 & 0.58 $\pm$ 0.04 \\
        %& DDPM~\cite{10.5555/3495724.3496298} & - $\pm$ - & - $\pm$ - & - $\pm$ - & - $\pm$ - \\
        & SuperpixSSL~\cite{10.1007/978-3-030-58526-6_45} & 93.18 $\pm$ 0.24 & 87.23 $\pm$ 0.42 & 1.76 $\pm$ 0.24 & 0.60 $\pm$ 0.03 \\
        & EM~\cite{8954439} & 93.04 $\pm$ 0.59 & 87.00 $\pm$ 1.02 & 1.73 $\pm$ 0.37 & 0.60 $\pm$ 0.09 \\
        & CCT~\cite{9157032} & 93.30 $\pm$ 0.20 & 87.45 $\pm$ 0.36 & 1.51 $\pm$ 0.28 & \textcolor{Blue}{\textbf{0.54 $\pm$ 0.05}} \\
        & UAMT~\cite{10.1007/978-3-030-32245-8_67} & \textcolor{Blue}{\textbf{93.42 $\pm$ 0.82}} & \textcolor{Blue}{\textbf{87.68 $\pm$ 1.55}} & 1.54 $\pm$ 0.46 & 0.56 $\pm$ 0.12 \\
        & CPS~\cite{9577639} & 93.04 $\pm$ 0.12 & 86.99 $\pm$ 0.22 & \textcolor{Blue}{\textbf{1.48 $\pm$ 0.16}} & 0.55 $\pm$ 0.02 \\
        & URPC~\cite{LUO2022102517} & 91.30 $\pm$ 1.44 & 84.05 $\pm$ 2.47 & 2.79 $\pm$ 1.28 & 0.88 $\pm$ 0.20 \\
        %& MT~\cite{10.5555/3294771.3294885} & 93.26 $\pm$ 1.20 & 88.97 $\pm$ 1.93 & 0.64 $\pm$ 0.36 & 0.09 $\pm$ 0.04 \\
        %& LHFC~\cite{10376766} & 96.08 $\pm$ 0.70 & 93.72 $\pm$ 0.76 & 0.03 $\pm$ 0.01 & 0.03 $\pm$ 0.00 \\
        & \cellcolor{GreenYellow} \textbf{Ours} & \textcolor{Red}{\textbf{93.82 $\pm$ 0.16}} & \textcolor{Red}{\textbf{88.35 $\pm$ 0.29}} & \textcolor{Red}{\textbf{1.26 $\pm$ 0.13}} & \textcolor{Red}{\textbf{0.46 $\pm$ 0.01}} \\
        \bottomrule
    \end{tabularx}
\end{table}

\paragraph{HMEPS} The outcomes on the HMEPS dataset are presented in Tab.~\ref{tab:comparison-sota-hmeps}. Again, our approach obtains improved SOTA performance in almost all the considered settings. However, we did not reach the best results with regime 1\%. We deem that this behavior can be linked to the intrinsic characteristics of this specific scenario, i.e., the HMEPS dataset provides a bigger set of labeled samples, even for the considered low training data regimes, making the unsupervised step less significant.

\begin{table}[tb]
    \caption{Ablation on the temperature hyperparameter. Mean $\pm$ 90\% CI are reported. The best results are in \textbf{bold}.}%
    \label{tab:ablation_temp}%
    \newcolumntype{L}{>{\centering\arraybackslash}m{.33\linewidth}}%
    \newcolumntype{M}{>{\centering\arraybackslash}m{.2\linewidth}}%
    \newcolumntype{C}{>{\centering\arraybackslash}X}
    \scriptsize%
    \centering
    \begin{tabularx}{0.9\linewidth}{L|M|C}
        \toprule
        Dataset (20\% Labeled) & Temperature & DC (\%) $\uparrow$ \\
        \midrule
        \multirow{7}{*}{GlaS~\cite{SIRINUKUNWATTANA2017489}} & 1 & 82.34 $\pm$ 1.06 \\
        & 5 & 82.90 $\pm$ 0.68 \\
        & 10 & 83.01 $\pm$ 0.68 \\
        & 20 & 83.39 $\pm$ 0.58 \\
        & 50 & 83.84 $\pm$ 0.71 \\
        & 75 & 84.15 $\pm$ 0.50 \\
        & \textbf{100} & \textbf{84.50 $\pm$ 0.50} \\
        \midrule
        \multirow{7}{*}{PH2~\cite{6610779}} & 1 & 86.57 $\pm$ 1.51 \\
        & 5 & 86.81 $\pm$ 1.16 \\
        & 10 & 89.00 $\pm$ 0.98 \\
        & \textbf{20} & \textbf{91.99 $\pm$ 0.82} \\
        & 50 & 88.48 $\pm$ 0.72 \\
        & 75 & 89.05 $\pm$ 1.06 \\
        & 100 & 89.15 $\pm$ 0.86 \\
        \midrule
        \multirow{7}{*}{HMEPS~\cite{raffaele_mazziotti_2021_4488164}} & 1 & 92.10 $\pm$ 0.38 \\
        & 5 & 92.84 $\pm$ 0.85 \\
        & 10 & 93.10 $\pm$ 0.94 \\
        & \textbf{20} & \textbf{93.82 $\pm$ 0.16} \\
        & 50 & 93.41 $\pm$ 0.23 \\
        & 75 & 93.38 $\pm$ 0.34 \\
        & 100 & 92.98 $\pm$ 0.85 \\
        \bottomrule
    \end{tabularx}
\end{table}

\begin{table}[!tb]
    \caption{Ablation on the type of Hebbian learning algorithm considering the GlaS dataset. The best results are in \textbf{bold}.}%
    \label{tab:ablation_hebb_formulations}%
    \newcolumntype{L}{>{\centering\arraybackslash}m{.22\linewidth}}%
    \newcolumntype{M}{>{\centering\arraybackslash}m{.31\linewidth}}%
    \newcolumntype{C}{>{\centering\arraybackslash}X}
    \scriptsize%
        %\captionsetup{font=footnotesize, labelformat=empty}
        %\caption{GlaS dataset~\cite{SIRINUKUNWATTANA2017489}}
        %\vspace*{-2mm}
        \centering
        \begin{tabularx}{0.999\linewidth}{L|M|C}
            \toprule
            Labeled\% & Method & DC (\%) $\uparrow$ \\
            \midrule
            \multirow{4}{*}{1\%} & HPCA-S & 66.18 $\pm$ 1.60 \\
            & HPCA-TSA & 65.18 $\pm$ 3.63 \\
            & SWTA-S & 68.26 $\pm$ 1.67 \\
            & SWTA-TSA & \textbf{69.95 $\pm$ 1.09} \\
            \midrule
            \multirow{4}{*}{2\%} & HPCA-S & 68.38 $\pm$ 1.77 \\
            & HPCA-TSA & 69.53 $\pm$ 1.27 \\
            & SWTA-S & 69.79 $\pm$ 1.20 \\
            & SWTA-TSA & \textbf{71.18 $\pm$ 1.28} \\
            \midrule
            \multirow{4}{*}{5\%} & HPCA-S & 75.19 $\pm$ 1.63 \\
            & HPCA-TSA & 73.62 $\pm$ 2.58 \\
            & SWTA-S & 74.89 $\pm$ 1.54 \\
            & SWTA-TSA & \textbf{77.18 $\pm$ 1.05} \\
            \midrule
            \multirow{4}{*}{10\%} & HPCA-S & 80.02 $\pm$ 1.23 \\
            & HPCA-TSA & 79.73 $\pm$ 1.12 \\
            & SWTA-S & 79.82 $\pm$ 1.06 \\
            & SWTA-TSA & \textbf{80.77 $\pm$ 0.77} \\
            \midrule
            \multirow{4}{*}{20\%} & HPCA-S & 83.20 $\pm$ 0.58 \\
            & HPCA-TSA & 83.61 $\pm$ 0.75 \\
            & SWTA-S & 84.00 $\pm$ 0.62 \\
            & SWTA-TSA &  \textbf{84.50 $\pm$ 0.50} \\
            \bottomrule
        \end{tabularx}
\end{table}

\begin{table}[!tb]
    \caption{Ablation of the first stage of our semi-supervised pipeline. The best results are in \textbf{bold}.}%
    \label{tab:ablation_first_stage}%
    \newcolumntype{C}{>{\centering\arraybackslash}X}
    \scriptsize%
        %\captionsetup{font=footnotesize, labelformat=empty}
        %\caption{GlaS dataset~\cite{SIRINUKUNWATTANA2017489}}
        %\vspace*{-2mm}
        \centering
        \begin{tabularx}{0.999\linewidth}{C|C}
            \toprule
            Method & DC (\%) $\uparrow$ \\
            \midrule
            VAE~\cite{DBLP:journals/corr/KingmaW13} & 57.5 \\
            SuperpixSSL~\cite{10.1007/978-3-030-58526-6_45} & 44.9 \\
            \textbf{Ours} & \textbf{59.4} \\
            \bottomrule
        \end{tabularx}
\end{table}

\begin{figure*}[!tb]
  \centering
  \begin{subfigure}{0.85\linewidth}
    \includegraphics[width=0.99\linewidth]{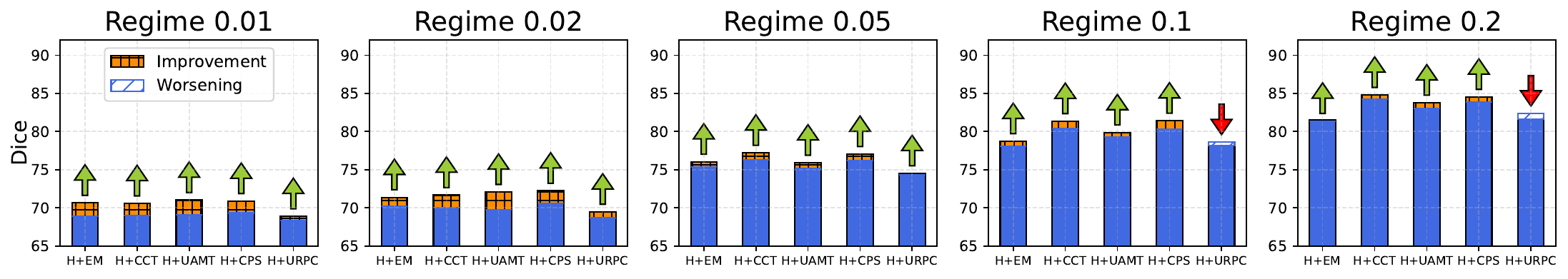}
    \caption{GlaS dataset~\cite{SIRINUKUNWATTANA2017489}}
    \label{fig:plots-glas}
  \end{subfigure}
  \begin{subfigure}{0.85\linewidth}
    \includegraphics[width=0.99\linewidth]{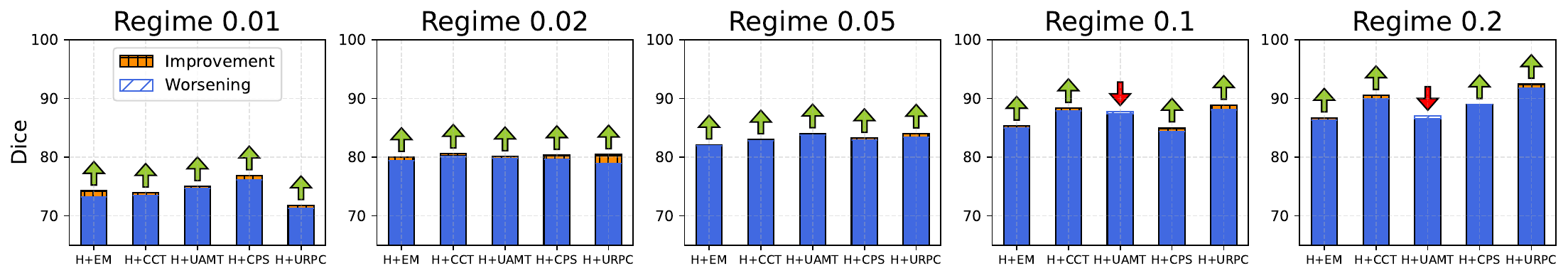}
    \caption{PH2 dataset~\cite{6610779}}
    \label{fig:plots-ph2}
  \end{subfigure}
  \begin{subfigure}{0.85\linewidth}
    \includegraphics[width=0.99\linewidth]{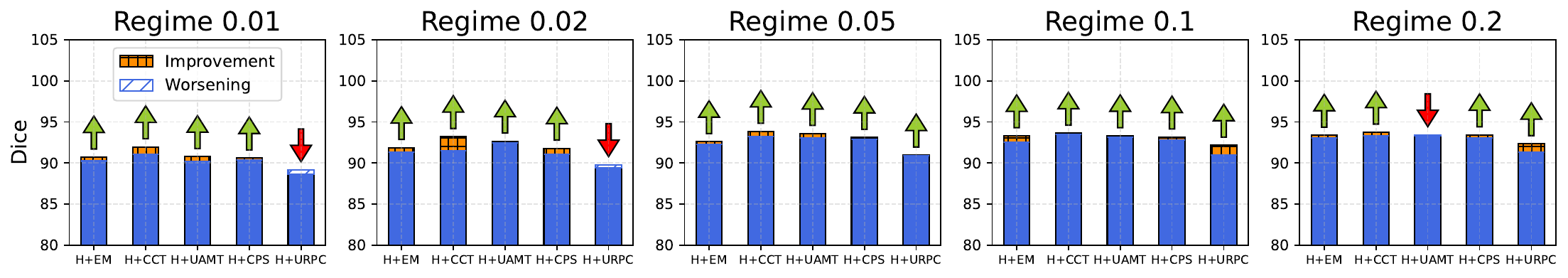}
    \caption{HMEPS dataset~\cite{raffaele_mazziotti_2021_4488164}}
    \label{fig:plots-hmeps}
  \end{subfigure}
  \caption{Performance changes (with green and red arrows) obtained with single-stage SOTA semi-supervised approaches initialized with our unsupervised Hebbian pre-training compared to initialization from scratch (blue bar). Each row corresponds to a dataset, while each column to a different degree of label availability.
  We report DC values embedded in the most convenient range for best readability.}
  \label{fig:plots}
\end{figure*}

\begin{table}[!tb]
    \centering
    \caption{Experiments with 3D images on the LA dataset~\cite{XIONG2021101832}. 
    %with 1\%, 2\%, 5\%, 10\%, and 20\% labeled data. 
    %Mean $\pm$ 90\% CI are reported. Best results are in red.
    }%
    \label{tab:comparison-sota-la}%
    \newcolumntype{L}{>{\centering\arraybackslash}m{.1\linewidth}}%
    \newcolumntype{M}{>{\centering\arraybackslash}m{.15\linewidth}}%
    \newcolumntype{C}{>{\centering\arraybackslash}m{.14\linewidth}}
    \tiny%
    \setlength{\tabcolsep}{3pt}
    \begin{tabularx}{0.98\linewidth}{L|M|CCCC}
        \toprule
        Labeled \% & Method & DC (\%) $\uparrow$ & JI (\%) $\uparrow$ & 95HD $\downarrow$ & ASD $\downarrow$ \\
        \midrule
        100\% & Fully Sup. & 91.76 $\pm$ 0.11 & 84.77 $\pm$ 0.19 & 5.75 $\pm$ 0.84 & 1.69 $\pm$ 0.06 \\
        \midrule
        \midrule
        %\multirow{6}{*}{1\%} & VAE~\cite{DBLP:journals/corr/KingmaW13} & - $\pm$ - & - $\pm$ - & - $\pm$ - & - $\pm$ - \\
        %& DDPM~\cite{10.5555/3495724.3496298} & - $\pm$ - & - $\pm$ - & - $\pm$ - & - $\pm$ - \\
        %& SuperpixSSL~\cite{10.1007/978-3-030-58526-6_45} & - $\pm$ - & - $\pm$ - & - $\pm$ - & - $\pm$ - \\
        \multirow{4}{*}{1\%} & EM~\cite{8954439} & 64.00 $\pm$ 3.55 & 47.09 $\pm$ 3.84 & 37.15 $\pm$ 4.89 & 9.55 $\pm$ 0.69 \\
        %& CCT~\cite{9157032} & 61.12 $\pm$ 6.87 & 44.27 $\pm$ 7.19 & 38.73 $\pm$ 6.41 & 9.65 $\pm$ 1.91 \\
        & UAMT~\cite{10.1007/978-3-030-32245-8_67} & 60.42 $\pm$ 4.56 & 43.39 $\pm$ 4.56 & 40.92 $\pm$ 5.49 & 10.37 $\pm$ 1.65 \\
        %& CPS~\cite{9577639} & 62.47 $\pm$ 7.96 & 45.80 $\pm$ 8.88 & 38.06 $\pm$ 8.94 & 8.79 $\pm$ 2.89 \\
        %& URPC~\cite{LUO2022102517} & - $\pm$ - & - $\pm$ - & - $\pm$ - & - $\pm$ - \\
        & DTC~\cite{Luo_2021} & \textcolor{Blue}{\textbf{62.63 $\pm$ 4.77}} & \textcolor{Blue}{\textbf{45.66 $\pm$ 4.99}} & \textcolor{Red}{\textbf{35.47 $\pm$ 3.34}} & \textcolor{Red}{\textbf{9.41 $\pm$ 0.61}} \\
        & \cellcolor{GreenYellow} Ours & \textcolor{Red}{\textbf{65.08 $\pm$ 4.74}} & \textcolor{Red}{\textbf{48.26 $\pm$ 4.54}} & \textcolor{Blue}{\textbf{40.82 $\pm$ 3.66}} & \textcolor{Blue}{\textbf{10.15 $\pm$ 1.41}} \\
        \midrule
        %
        %\multirow{6}{*}{2\%} & VAE~\cite{DBLP:journals/corr/KingmaW13} & - $\pm$ - & - $\pm$ - & - $\pm$ - & - $\pm$ - \\
        %& DDPM~\cite{10.5555/3495724.3496298} & - $\pm$ - & - $\pm$ - & - $\pm$ - & - $\pm$ - \\
        %& SuperpixSSL~\cite{10.1007/978-3-030-58526-6_45} & - $\pm$ - & - $\pm$ - & - $\pm$ - & - $\pm$ - \\
        \multirow{4}{*}{2\%} & EM~\cite{8954439} & 73.53 $\pm$ 4.09 & 58.36 $\pm$ 6.96 & 31.35 $\pm$ 6.85 & 7.50 $\pm$ 1.72 \\
        %& CCT~\cite{9157032} & 75.49 $\pm$ 2.67 & 60.63 $\pm$ 3.45 & 31.00 $\pm$ 29.08 & 6.53 $\pm$ 3.52 \\
        & UAMT~\cite{10.1007/978-3-030-32245-8_67} & 74.80 $\pm$ 5.18 & 59.93 $\pm$ 6.42 & 28.05 $\pm$ 8.61 & 6.52 $\pm$ 2.18 \\
        %& CPS~\cite{9577639} & - $\pm$ - & - $\pm$ - & - $\pm$ - & - $\pm$ - \\
        %& URPC~\cite{LUO2022102517} & - $\pm$ - & - $\pm$ - & - $\pm$ - & - $\pm$ - \\
        & DTC~\cite{Luo_2021} & \textcolor{Red}{\textbf{76.87 $\pm$ 4.50}} & \textcolor{Red}{\textbf{62.51 $\pm$ 5.88}} & \textcolor{Blue}{\textbf{25.74 $\pm$ 2.94}} & \textcolor{Blue}{\textbf{5.82 $\pm$ 0.74}} \\
        & \cellcolor{GreenYellow} Ours & \textcolor{Blue}{\textbf{76.42 $\pm$ 1.87}} & \textcolor{Blue}{\textbf{61.86 $\pm$ 2.46}} & \textcolor{Red}{\textbf{24.66 $\pm$ 2.80}} & \textcolor{Red}{\textbf{5.73 $\pm$ 0.43}} \\
        \midrule
        %
        %\multirow{6}{*}{5\%} & VAE~\cite{DBLP:journals/corr/KingmaW13} & - $\pm$ - & - $\pm$ - & - $\pm$ - & - $\pm$ - \\
        %& DDPM~\cite{10.5555/3495724.3496298} & - $\pm$ - & - $\pm$ - & - $\pm$ - & - $\pm$ - \\
        %& SuperpixSSL~\cite{10.1007/978-3-030-58526-6_45} & - $\pm$ - & - $\pm$ - & - $\pm$ - & - $\pm$ - \\
        \multirow{4}{*}{5\%} & EM~\cite{8954439} & 80.80 $\pm$ 3.73 & 67.84 $\pm$ 5.33 & 23.12 $\pm$ 5.78 & 5.40 $\pm$ 1.17 \\
        %& CCT~\cite{9157032} & \textcolor{Red}{\underline{\textbf{84.70 $\pm$ 0.27}}} & \textcolor{Red}{\underline{\textbf{73.47 $\pm$ 0.42}}} & 17.58 $\pm$ 4.21 & 3.89 $\pm$ 0.73 \\
        & UAMT~\cite{10.1007/978-3-030-32245-8_67} & \textcolor{Red}{\textbf{84.22 $\pm$ 3.67}} & \textcolor{Red}{\textbf{72.80 $\pm$ 5.41}} & \textcolor{Red}{\textbf{15.97 $\pm$ 2.28}} & \textcolor{Red}{\textbf{3.73 $\pm$ 0.42}} \\
        %& CPS~\cite{9577639} & - $\pm$ - & - $\pm$ - & - $\pm$ - & - $\pm$ - \\
        %& URPC~\cite{LUO2022102517} & - $\pm$ - & - $\pm$ - & - $\pm$ - & - $\pm$ - \\
        & DTC~\cite{Luo_2021} & 83.28 $\pm$ 2.97 & 71.43 $\pm$ 4.33 & 17.55 $\pm$ 3.80 & 4.19 $\pm$ 0.94 \\
        & \cellcolor{GreenYellow} Ours & \textcolor{Blue}{\textbf{83.42 $\pm$ 1.92}} & \textcolor{Blue}{\textbf{71.57 $\pm$ 2.82}} & \textcolor{Blue}{\textbf{17.40 $\pm$ 5.38}} & \textcolor{Blue}{\textbf{4.18 $\pm$ 0.60}} \\
        \midrule
        %
        %\multirow{6}{*}{10\%} & VAE~\cite{DBLP:journals/corr/KingmaW13} & - $\pm$ - & - $\pm$ - & - $\pm$ - & - $\pm$ - \\
        %& DDPM~\cite{10.5555/3495724.3496298} & - $\pm$ - & - $\pm$ - & - $\pm$ - & - $\pm$ - \\
        %& SuperpixSSL~\cite{10.1007/978-3-030-58526-6_45} & - $\pm$ - & - $\pm$ - & - $\pm$ - & - $\pm$ - \\
        \multirow{4}{*}{10\%} & EM~\cite{8954439} & 85.33 $\pm$ 1.42 & 74.41 $\pm$ 2.18 & 17.72 $\pm$ 0.98 & 3.79 $\pm$ 0.44 \\
        %& CCT~\cite{9157032} & 88.11 $\pm$ 1.18 & 78.75 $\pm$ 1.89 & 13.26 $\pm$ 1.30 & - $\pm$ - \\
        & UAMT~\cite{10.1007/978-3-030-32245-8_67} & \textcolor{Red}{\textbf{88.95 $\pm$ 0.22}} & \textcolor{Red}{\textbf{80.09 $\pm$ 0.37}} & \textcolor{Red}{\textbf{8.13 $\pm$ 0.02}} & \textcolor{Red}{\textbf{2.21 $\pm$ 0.06}} \\
        %& CPS~\cite{9577639} & - $\pm$ - & - $\pm$ - & - $\pm$ - & - $\pm$ - \\
        %& URPC~\cite{LUO2022102517} & - $\pm$ - & - $\pm$ - & - $\pm$ - & - $\pm$ - \\
        & DTC~\cite{Luo_2021} & 86.43 $\pm$ 1.73 & 76.13 $\pm$ 2.69 & 13.71 $\pm$ 4.95 & 3.22 $\pm$ 0.74 \\
        & \cellcolor{GreenYellow} Ours & \textcolor{Blue}{\textbf{87.11 $\pm$ 0.69}} & \textcolor{Blue}{\textbf{77.17 $\pm$ 1.07}} & \textcolor{Blue}{\textbf{13.63 $\pm$ 2.16}} & \textcolor{Blue}{\textbf{3.21 $\pm$ 0.11}} \\
        \midrule
        %
        %\multirow{6}{*}{20\%} & VAE~\cite{DBLP:journals/corr/KingmaW13} & - $\pm$ - & - $\pm$ - & - $\pm$ - & - $\pm$ - \\
        %& DDPM~\cite{10.5555/3495724.3496298} & - $\pm$ - & - $\pm$ - & - $\pm$ - & - $\pm$ - \\
        %& SuperpixSSL~\cite{10.1007/978-3-030-58526-6_45} & - $\pm$ - & - $\pm$ - & - $\pm$ - & - $\pm$ - \\
        \multirow{4}{*}{20\%} & EM~\cite{8954439} & \textcolor{Blue}{\textbf{89.51 $\pm$ 0.26}} & \textcolor{Blue}{\textbf{81.01 $\pm$ 0.42}} & 9.61 $\pm$ 1.34 & 2.43 $\pm$ 0.23 \\
        %& CCT~\cite{9157032} & - $\pm$ - & - $\pm$ - & - $\pm$ - & - $\pm$ - \\
        & UAMT~\cite{10.1007/978-3-030-32245-8_67} & \textcolor{Red}{\textbf{90.91 $\pm$ 0.64}} & \textcolor{Red}{\textbf{83.34 $\pm$ 1.07}} & \textcolor{Red}{\textbf{6.09 $\pm$ 1.00}} & \textcolor{Red}{\textbf{1.82 $\pm$ 0.04}} \\
        %& CPS~\cite{9577639} & - $\pm$ - & - $\pm$ - & - $\pm$ - & - $\pm$ - \\
        %& URPC~\cite{LUO2022102517} & - $\pm$ - & - $\pm$ - & - $\pm$ - & - $\pm$ - \\
        & DTC~\cite{Luo_2021} & 89.46 $\pm$ 1.45 & 80.93 $\pm$ 2.36 & \textcolor{Blue}{\textbf{8.75 $\pm$ 2.29}} & \textcolor{Blue}{\textbf{2.25 $\pm$ 0.11}} \\
        & \cellcolor{GreenYellow} Ours & 89.17 $\pm$ 1.25 & 80.45 $\pm$ 2.05 & 11.92 $\pm$ 7.96 & 2.66 $\pm$ 0.49 \\
        \midrule
        \multicolumn{6}{c}{{\includegraphics[align=c,height=1.5cm,width=7cm]{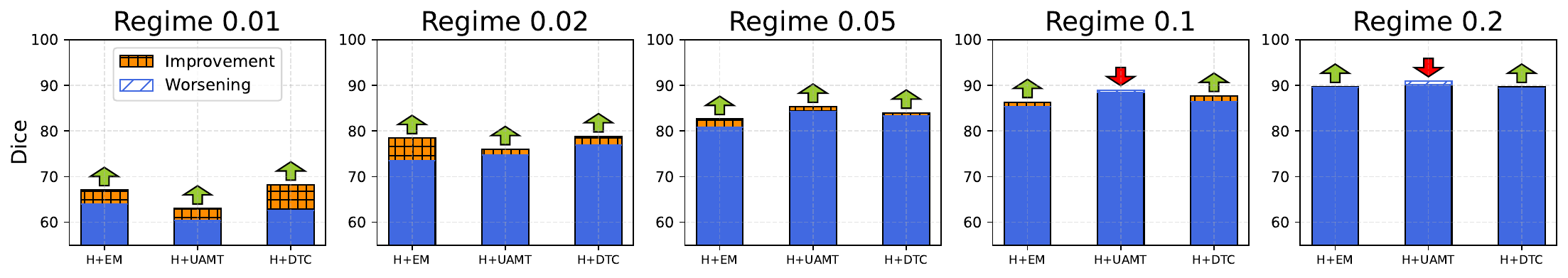}}} \\
        \bottomrule
    \end{tabularx}
\end{table}

\subsection{Hebbian Initialization of Single-stage Methods}
\label{sec:sec:additional-exp}

\noindent We report the results obtained by using our unsupervised SWTA-TSA pre-training as initialization for the considered single-stage pseudo-labeling and consistency-based semi-supervised methods.
Results are shown in Fig.~\ref{fig:plots}. For better readability, we illustrate only the performance in terms of the gold standard metric, i.e., the DC. We can observe that, in most cases, the proposed initialization helps achieve significant improvements. However, we can still notice a small worsening in some cases. We deem that this can be expected because a given initialization, although effective for some methods and specific data distributions/data regimes, can be sub-optimal for other techniques, which instead require starting with small random weights to guarantee the stability and convergence of the training process. 

\begin{figure*}[!t]
    \setlength{\fboxsep}{0pt}%
    \setlength{\fboxrule}{1pt}%
    \centering
    \newcolumntype{C}{>{\centering\arraybackslash}X}
    \setlength{\tabcolsep}{1pt}
    \begin{tabularx}{\linewidth}{l@{\hspace{5pt}}CCC|CCC|CCC}
        & \multicolumn{3}{c}{PH2~\cite{6610779}}
        & \multicolumn{3}{c}{GlaS~\cite{SIRINUKUNWATTANA2017489}}
        & \multicolumn{3}{c}{HMEPS~\cite{raffaele_mazziotti_2021_4488164}} \\
        \cmidrule(lr){2-4} \cmidrule(lr){5-7} \cmidrule(lr){8-10}
        & Sample
        & Target
        & Pred.
        & Sample
        & Target
        & Pred.
        & Sample
        & Target
        & Pred. \\
        \rotatebox[origin=c]{90}{\scriptsize Labeled 1\%}&%
        \fcolorbox{lime}{white}{\includegraphics[align=c,width=\linewidth,height=1.3cm]{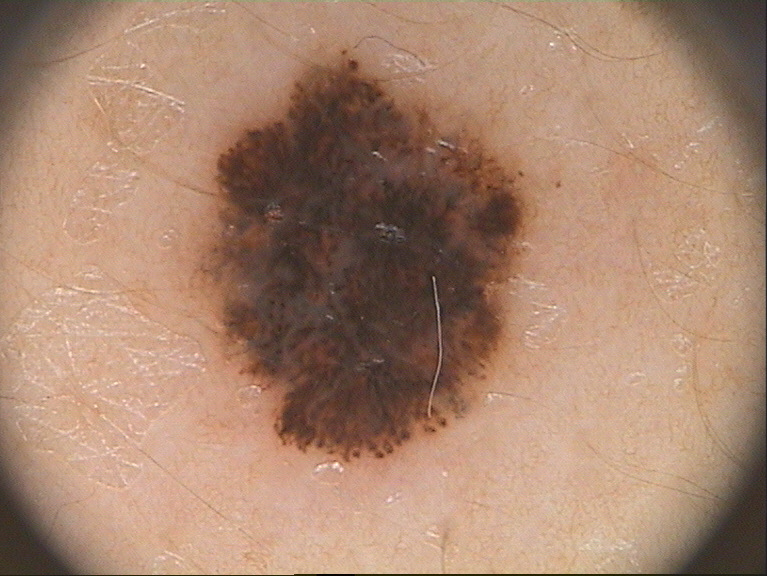}}&%
        \fcolorbox{magenta}{white}{\includegraphics[align=c,width=\linewidth,height=1.3cm]{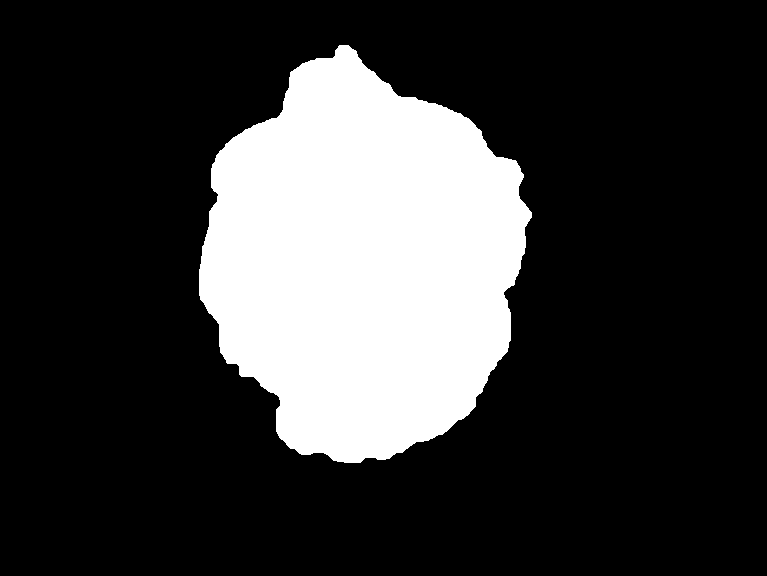}}&%
        \fcolorbox{cyan}{white}{\includegraphics[align=c,width=\linewidth,height=1.3cm]{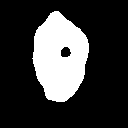}}&%
        \fcolorbox{lime}{white}{\includegraphics[align=c,width=\linewidth,height=1.3cm]{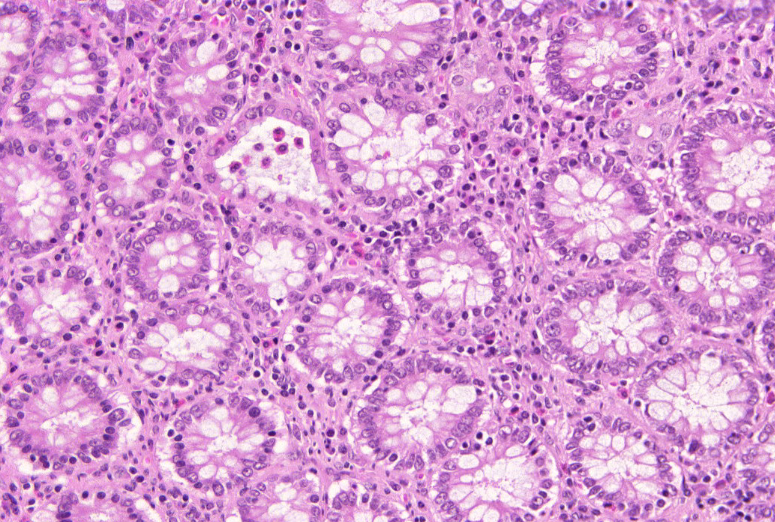}}&%
        \fcolorbox{magenta}{white}{\includegraphics[align=c,width=\linewidth,height=1.3cm]{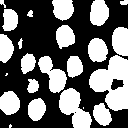}}&%
        \fcolorbox{cyan}{white}{\includegraphics[align=c,width=\linewidth,height=1.3cm]{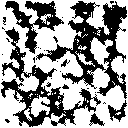}}&%
        \fcolorbox{lime}{white}{\includegraphics[align=c,width=\linewidth,height=1.3cm]{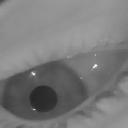}}&%
        \fcolorbox{magenta}{white}{\includegraphics[align=c,width=\linewidth,height=1.3cm]{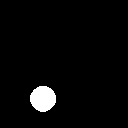}}&%
        \fcolorbox{cyan}{white}{\includegraphics[align=c,width=\linewidth,height=1.3cm]{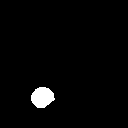}} \vspace{0.5ex}\\%
        \rotatebox[origin=c]{90}{\scriptsize Labeled 2\%}&%
        \fcolorbox{lime}{white}{\includegraphics[align=c,width=\linewidth,height=1.3cm]{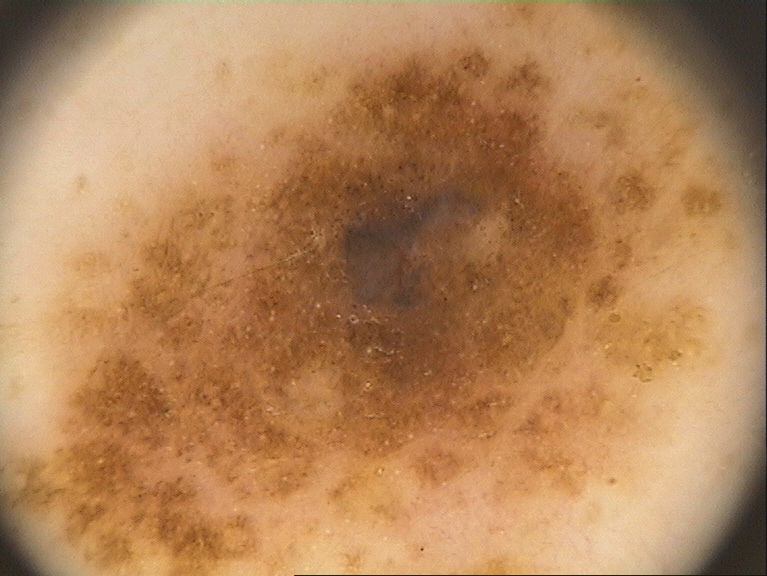}}&%
        \fcolorbox{magenta}{white}{\includegraphics[align=c,width=\linewidth,height=1.3cm]{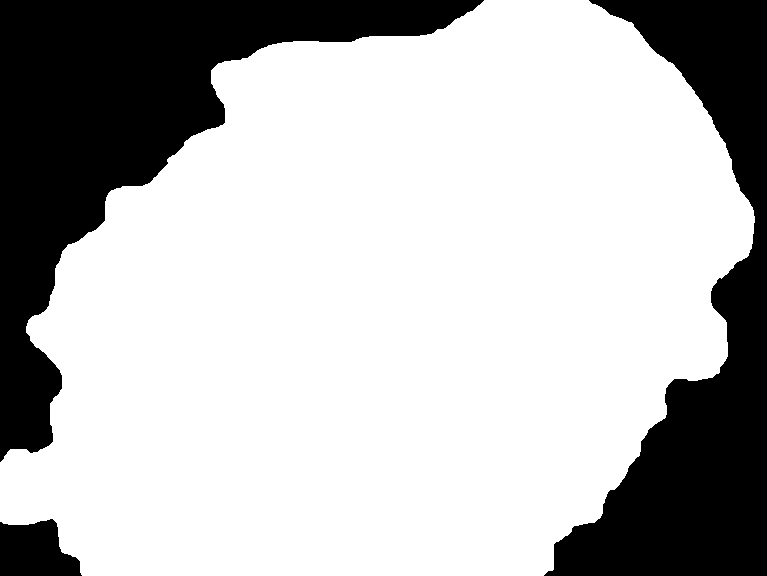}}&%
        \fcolorbox{cyan}{white}{\includegraphics[align=c,width=\linewidth,height=1.3cm]{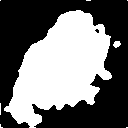}}&%
        \fcolorbox{lime}{white}{\includegraphics[align=c,width=\linewidth,height=1.3cm]{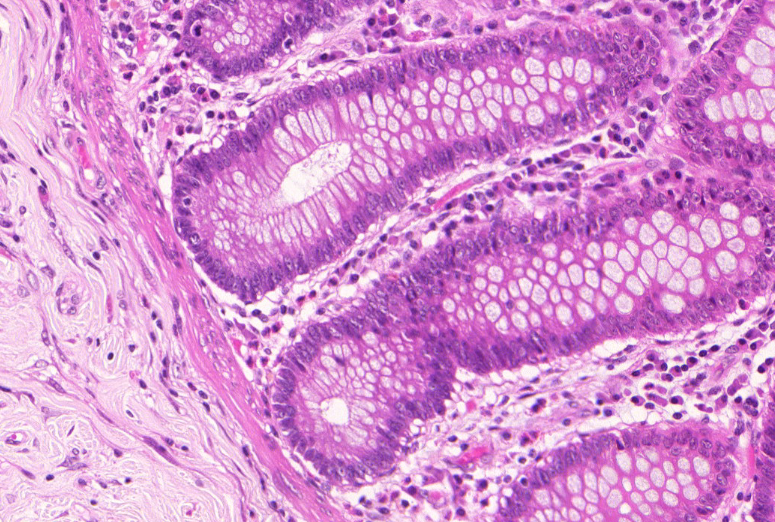}}&%
        \fcolorbox{magenta}{white}{\includegraphics[align=c,width=\linewidth,height=1.3cm]{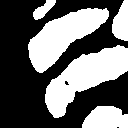}}&%
        \fcolorbox{cyan}{white}{\includegraphics[align=c,width=\linewidth,height=1.3cm]{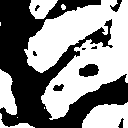}}&%
        \fcolorbox{lime}{white}{\includegraphics[align=c,width=\linewidth,height=1.3cm]{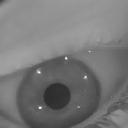}}&%
        \fcolorbox{magenta}{white}{\includegraphics[align=c,width=\linewidth,height=1.3cm]{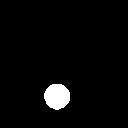}}&%
        \fcolorbox{cyan}{white}{\includegraphics[align=c,width=\linewidth,height=1.3cm]{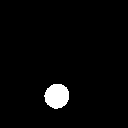}} \vspace{0.5ex}\\%
        \rotatebox[origin=c]{90}{\scriptsize Labeled 5\%}&%
        \fcolorbox{lime}{white}{\includegraphics[align=c,width=\linewidth,height=1.3cm]{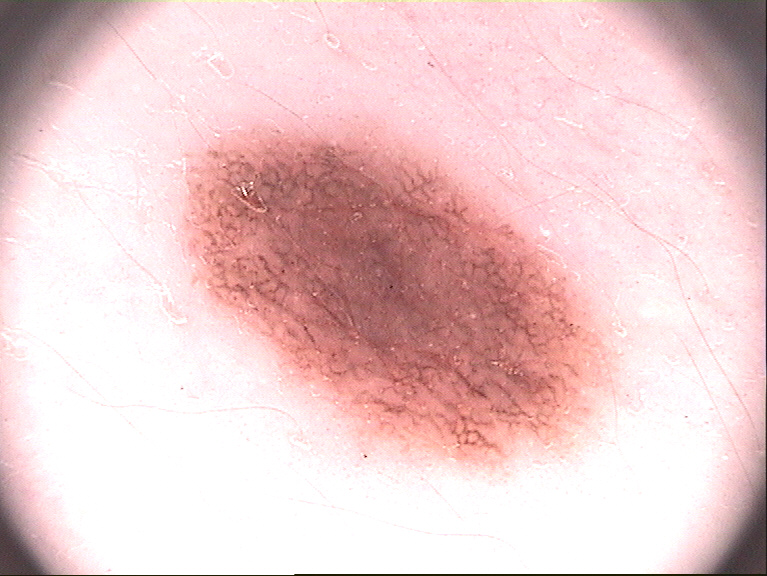}}&%
        \fcolorbox{magenta}{white}{\includegraphics[align=c,width=\linewidth,height=1.3cm]{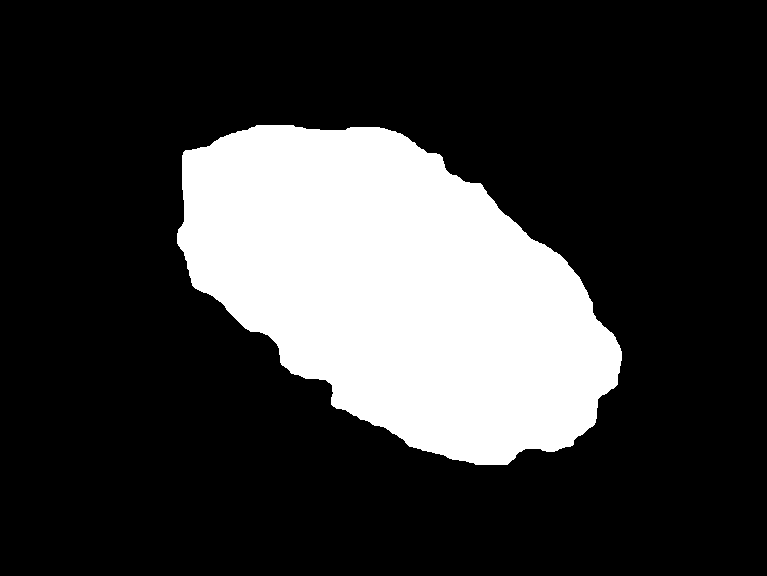}}&%
        \fcolorbox{cyan}{white}{\includegraphics[align=c,width=\linewidth,height=1.3cm]{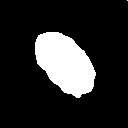}}&%
        \fcolorbox{lime}{white}{\includegraphics[align=c,width=\linewidth,height=1.3cm]{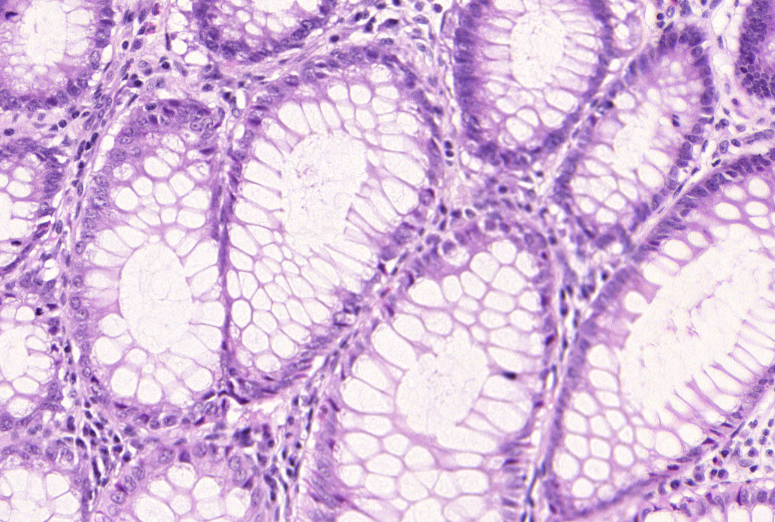}}&%
        \fcolorbox{magenta}{white}{\includegraphics[align=c,width=\linewidth,height=1.3cm]{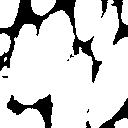}}&%
        \fcolorbox{cyan}{white}{\includegraphics[align=c,width=\linewidth,height=1.3cm]{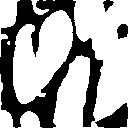}}&%
        \fcolorbox{lime}{white}{\includegraphics[align=c,width=\linewidth,height=1.3cm]{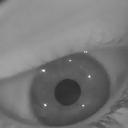}}&%
        \fcolorbox{magenta}{white}{\includegraphics[align=c,width=\linewidth,height=1.3cm]{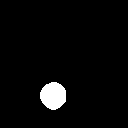}}&%
        \fcolorbox{cyan}{white}{\includegraphics[align=c,width=\linewidth,height=1.3cm]{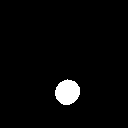}} \vspace{0.5ex}\\%
        \rotatebox[origin=c]{90}{\scriptsize Labeled 10\%}&%
        \fcolorbox{lime}{white}{\includegraphics[align=c,width=\linewidth,height=1.3cm]{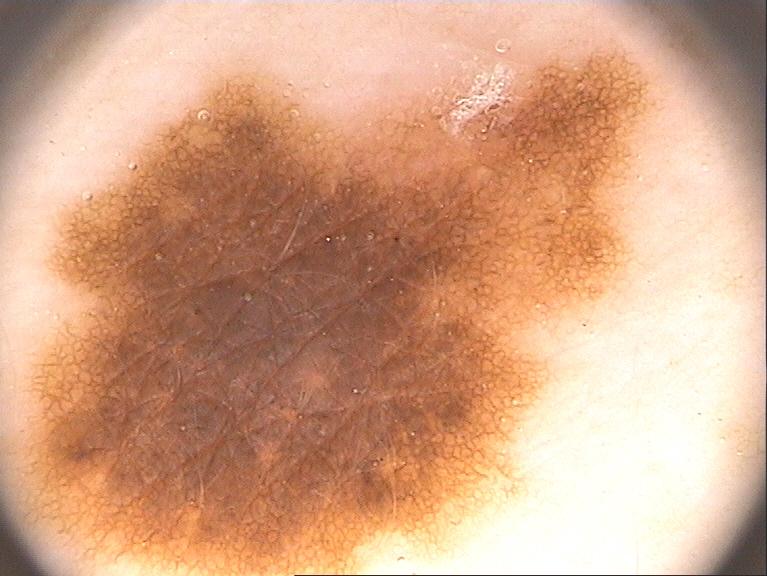}}&%
        \fcolorbox{magenta}{white}{\includegraphics[align=c,width=\linewidth,height=1.3cm]{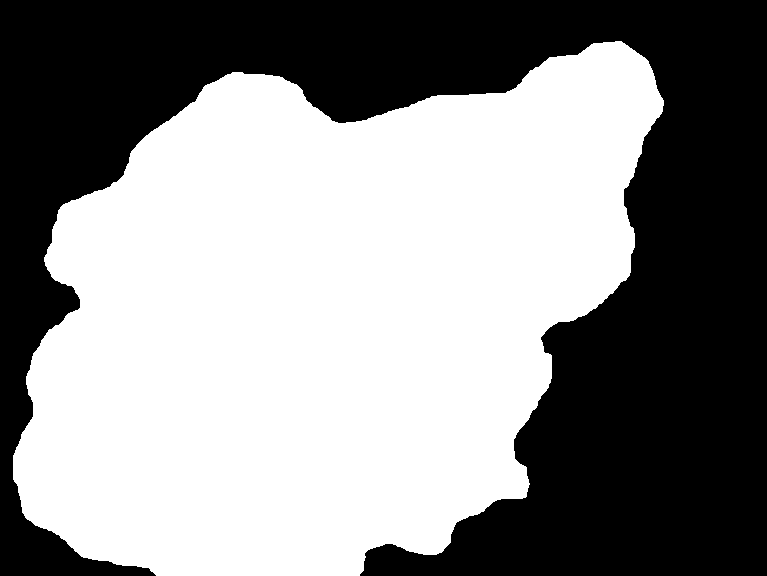}}&%
        \fcolorbox{cyan}{white}{\includegraphics[align=c,width=\linewidth,height=1.3cm]{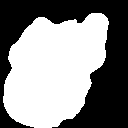}}&%
        \fcolorbox{lime}{white}{\includegraphics[align=c,width=\linewidth,height=1.3cm]{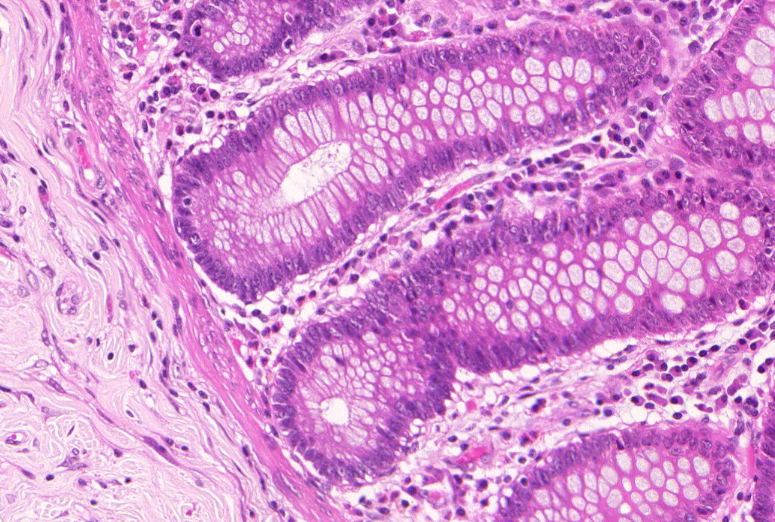}}&%
        \fcolorbox{magenta}{white}{\includegraphics[align=c,width=\linewidth,height=1.3cm]{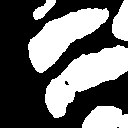}}&%
        \fcolorbox{cyan}{white}{\includegraphics[align=c,width=\linewidth,height=1.3cm]{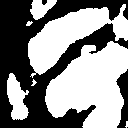}}&%
        \fcolorbox{lime}{white}{\includegraphics[align=c,width=\linewidth,height=1.3cm]{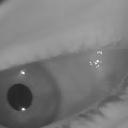}}&%
        \fcolorbox{magenta}{white}{\includegraphics[align=c,width=\linewidth,height=1.3cm]{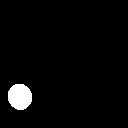}}&%
        \fcolorbox{cyan}{white}{\includegraphics[align=c,width=\linewidth,height=1.3cm]{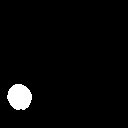}} \vspace{0.5ex}\\%
        \rotatebox[origin=c]{90}{\scriptsize Labeled 20\%}&%
        \fcolorbox{lime}{white}{\includegraphics[align=c,width=\linewidth,height=1.3cm]{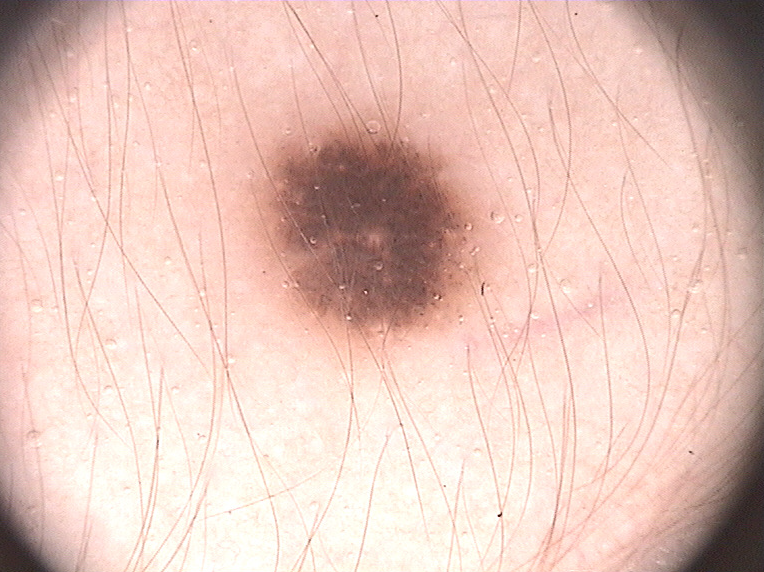}}&%
        \fcolorbox{magenta}{white}{\includegraphics[align=c,width=\linewidth,height=1.3cm]{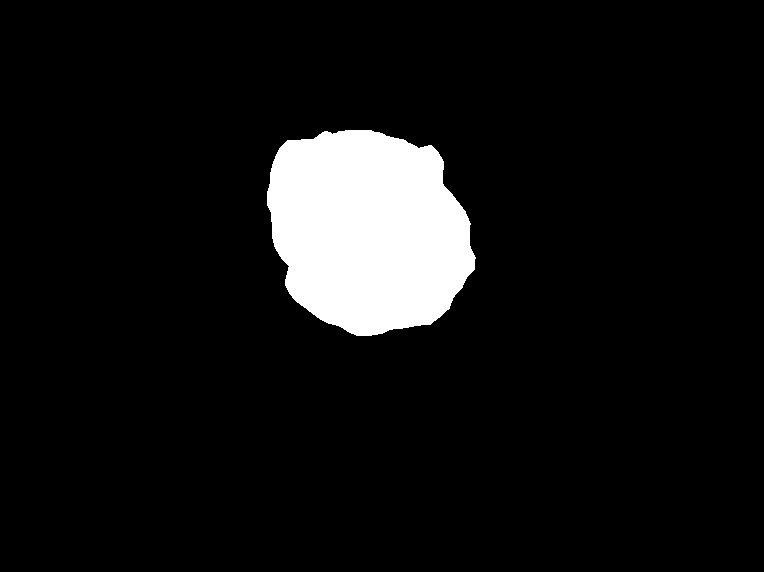}}&%
        \fcolorbox{cyan}{white}{\includegraphics[align=c,width=\linewidth,height=1.3cm]{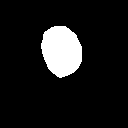}}&%
        \fcolorbox{lime}{white}{\includegraphics[align=c,width=\linewidth,height=1.3cm]{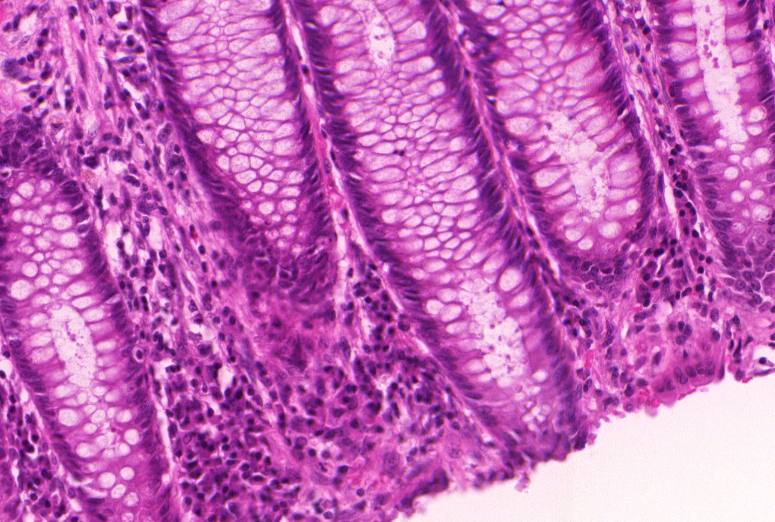}}&%
        \fcolorbox{magenta}{white}{\includegraphics[align=c,width=\linewidth,height=1.3cm]{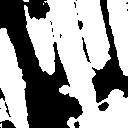}}&%
        \fcolorbox{cyan}{white}{\includegraphics[align=c,width=\linewidth,height=1.3cm]{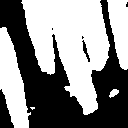}}&%
        \fcolorbox{lime}{white}{\includegraphics[align=c,width=\linewidth,height=1.3cm]{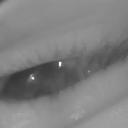}}&%
        \fcolorbox{magenta}{white}{\includegraphics[align=c,width=\linewidth,height=1.3cm]{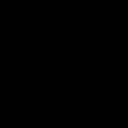}}&%
        \fcolorbox{cyan}{white}{\includegraphics[align=c,width=\linewidth,height=1.3cm]{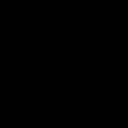}} \vspace{0.5ex}\\%
    \end{tabularx}
    \caption{Qualitative results from our semi-supervised approach based on Hebbian SWTA-TSA. Each row corresponds to a different percentage of label availability; each column corresponds to a different dataset and includes a triplet sample-target-prediction.}
    \label{fig:qualitatite-results}    
\end{figure*}

\begin{figure*}[!t]
    \setlength{\fboxsep}{0pt}%
    \setlength{\fboxrule}{1pt}%
    \centering
    \newcolumntype{C}{>{\centering\arraybackslash}X}
    \setlength{\tabcolsep}{1pt}
    \begin{tabularx}{\linewidth}{l@{\hspace{5pt}}CCC|CCC}
        & \multicolumn{3}{c}{LA~\cite{XIONG2021101832} (Slice view)}
        & \multicolumn{3}{c}{LA~\cite{XIONG2021101832} (3D view)} \\
        \cmidrule(lr){2-4} \cmidrule(lr){5-7}
        & Sample
        & Target
        & Pred.
        & Sample
        & Target
        & Pred. \\
        
        \rotatebox[origin=c]{90}{\scriptsize Labeled 1\%}&%
        \fcolorbox{lime}{white}{\includegraphics[align=c,width=\linewidth,height=1.3cm]{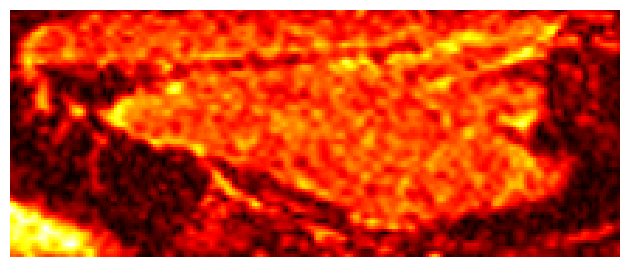}}&%
        \fcolorbox{magenta}{white}{\includegraphics[align=c,width=\linewidth,height=1.3cm]{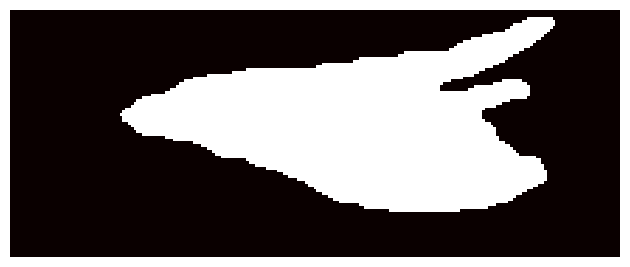}}&%
        \fcolorbox{cyan}{white}{\includegraphics[align=c,width=\linewidth,height=1.3cm]{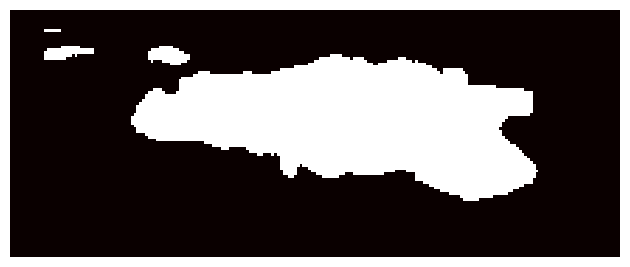}}&%
        \fcolorbox{lime}{white}{\includegraphics[align=c,width=\linewidth,height=1.3cm]{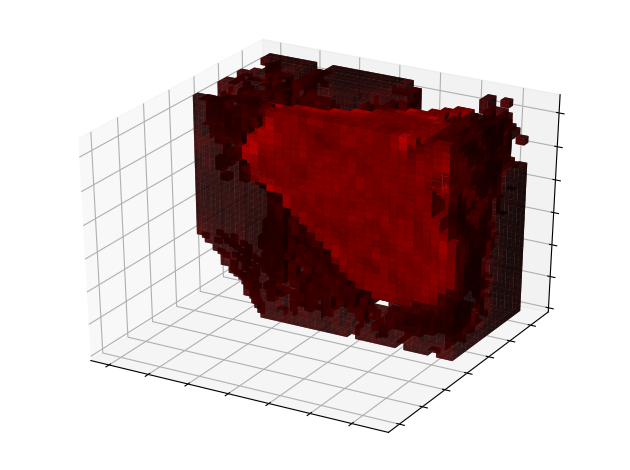}}&%
        \fcolorbox{magenta}{white}{\includegraphics[align=c,width=\linewidth,height=1.3cm]{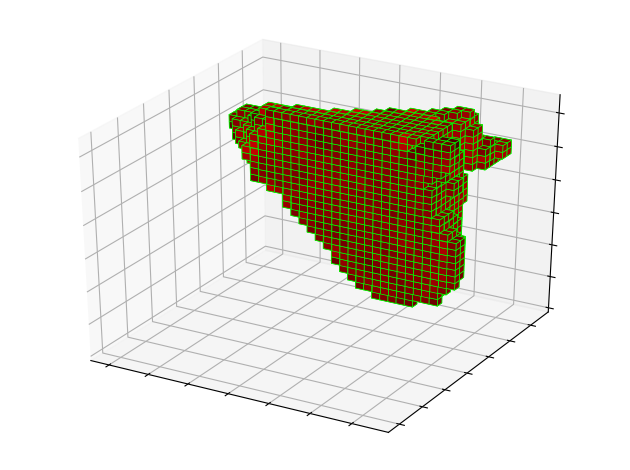}}&%
        \fcolorbox{cyan}{white}{\includegraphics[align=c,width=\linewidth,height=1.3cm]{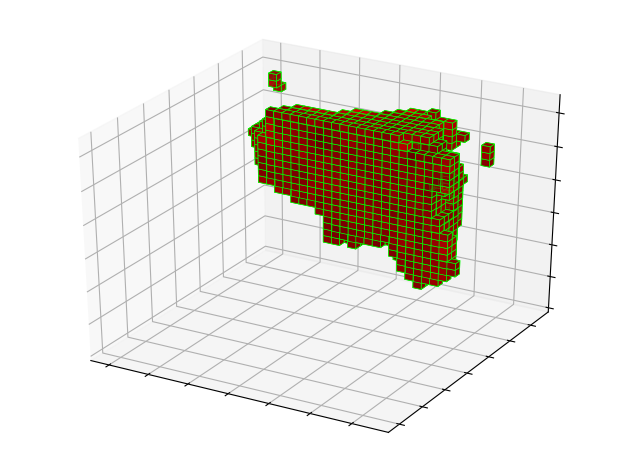}}\\%

        \rotatebox[origin=c]{90}{\scriptsize Labeled 2\%}&%
        \fcolorbox{lime}{white}
        {\includegraphics[align=c,width=\linewidth,height=1.3cm]{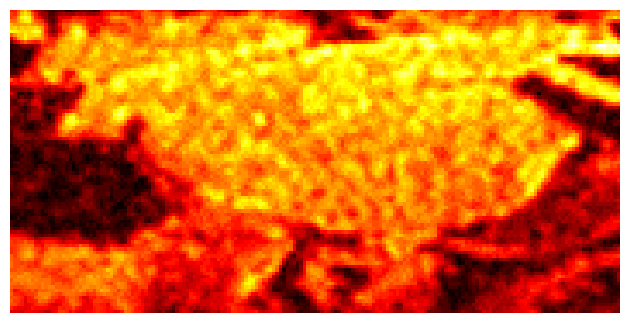}}&%
        \fcolorbox{magenta}{white}{\includegraphics[align=c,width=\linewidth,height=1.3cm]{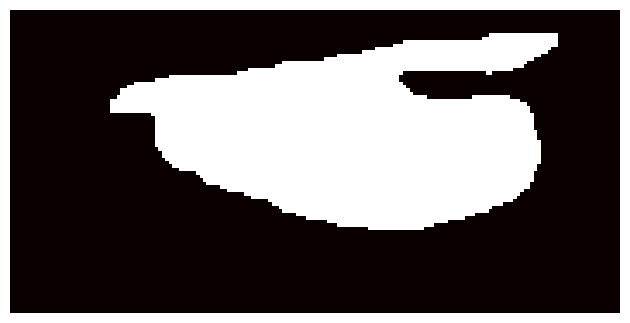}}&%
        \fcolorbox{cyan}{white}{\includegraphics[align=c,width=\linewidth,height=1.3cm]{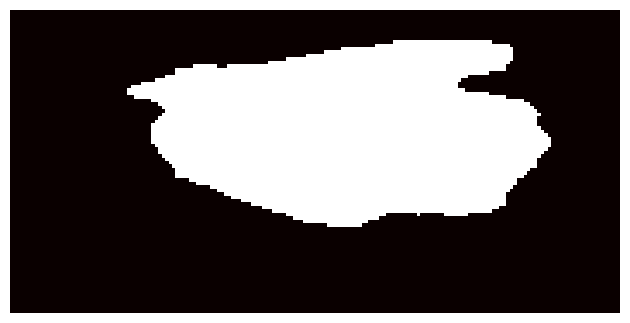}}&%
        \fcolorbox{lime}{white}{\includegraphics[align=c,width=\linewidth,height=1.3cm]{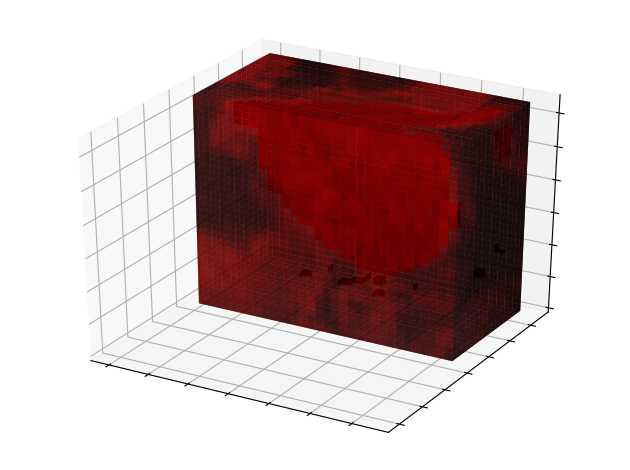}}&%
        \fcolorbox{magenta}{white}{\includegraphics[align=c,width=\linewidth,height=1.3cm]{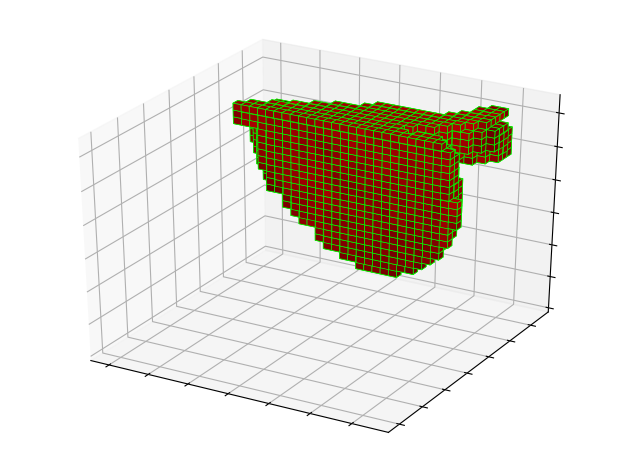}}&%
        \fcolorbox{cyan}{white}{\includegraphics[align=c,width=\linewidth,height=1.3cm]{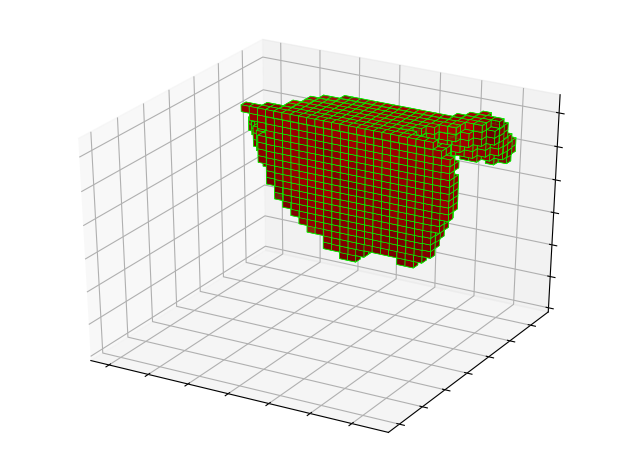}}\\%
        
        \rotatebox[origin=c]{90}{\scriptsize Labeled 5\%}&%
        \fcolorbox{lime}{white}{\includegraphics[align=c,width=\linewidth,height=1.3cm]{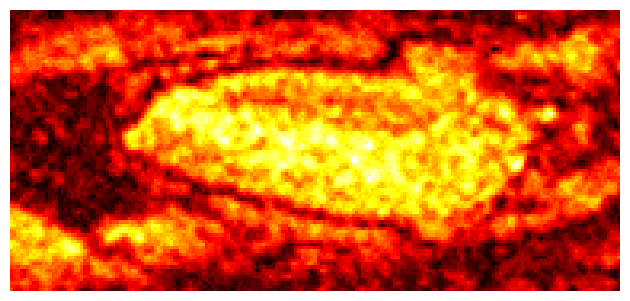}}&%
        \fcolorbox{magenta}{white}{\includegraphics[align=c,width=\linewidth,height=1.3cm]{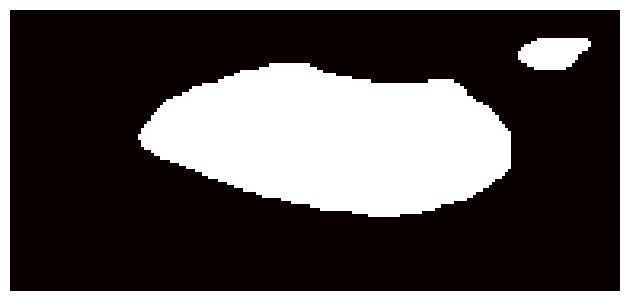}}&%
        \fcolorbox{cyan}{white}{\includegraphics[align=c,width=\linewidth,height=1.3cm]{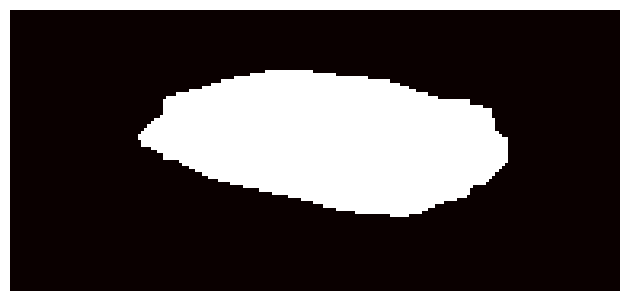}}&%
        \fcolorbox{lime}{white}{\includegraphics[align=c,width=\linewidth,height=1.3cm]{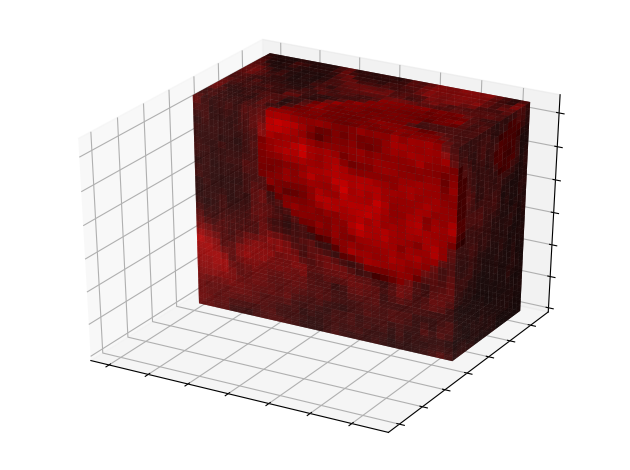}}&%
        \fcolorbox{magenta}{white}{\includegraphics[align=c,width=\linewidth,height=1.3cm]{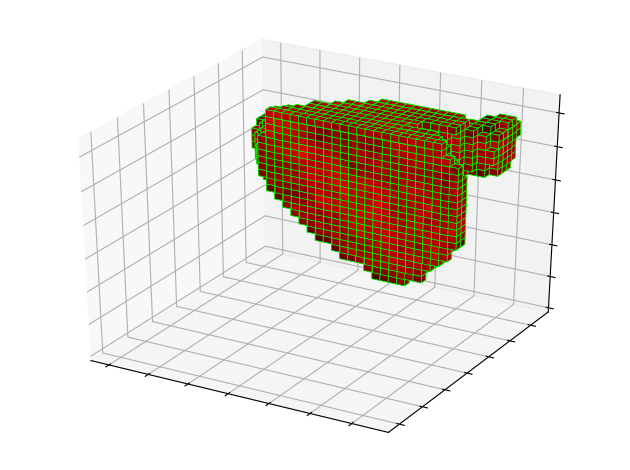}}&%
        \fcolorbox{cyan}{white}{\includegraphics[align=c,width=\linewidth,height=1.3cm]{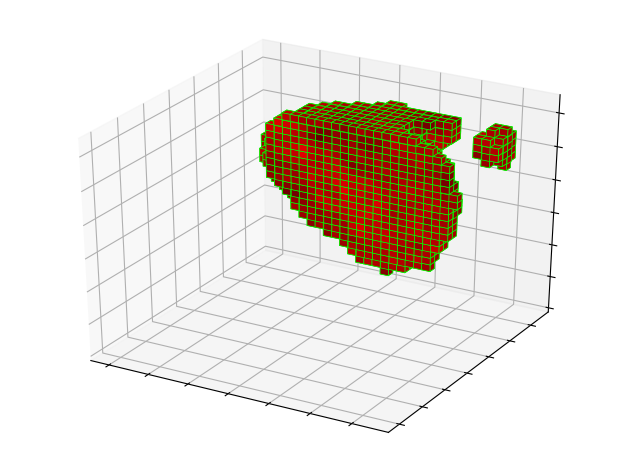}}\\%
        
        \rotatebox[origin=c]{90}{\scriptsize Labeled 10\%}&%
        \fcolorbox{lime}{white}{\includegraphics[align=c,width=\linewidth,height=1.3cm]{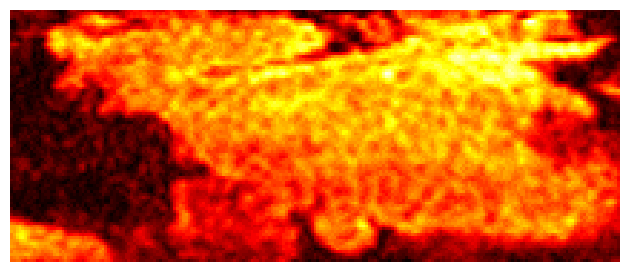}}&%
        \fcolorbox{magenta}{white}{\includegraphics[align=c,width=\linewidth,height=1.3cm]{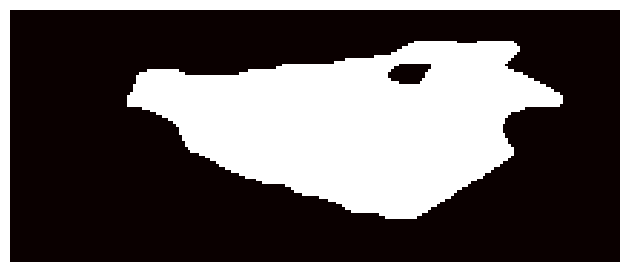}}&%
        \fcolorbox{cyan}{white}{\includegraphics[align=c,width=\linewidth,height=1.3cm]{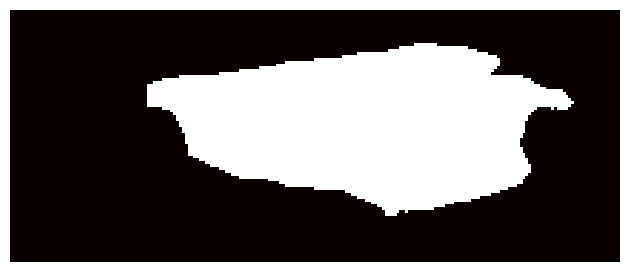}}&%
        \fcolorbox{lime}{white}{\includegraphics[align=c,width=\linewidth,height=1.3cm]{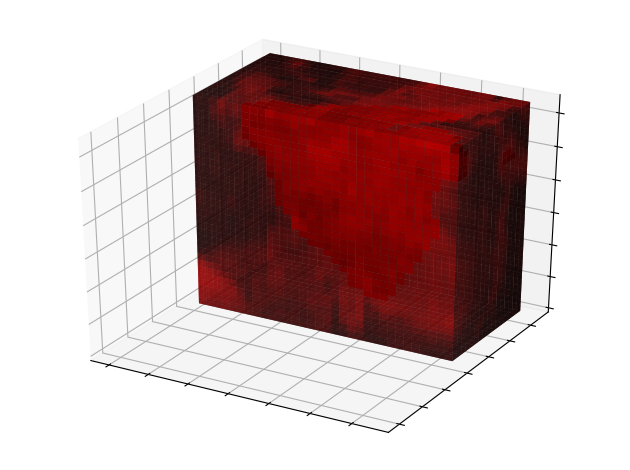}}&%
        \fcolorbox{magenta}{white}{\includegraphics[align=c,width=\linewidth,height=1.3cm]{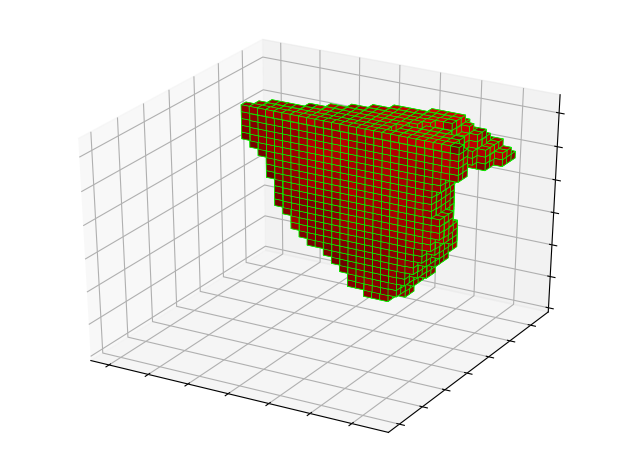}}&%
        \fcolorbox{cyan}{white}{\includegraphics[align=c,width=\linewidth,height=1.3cm]{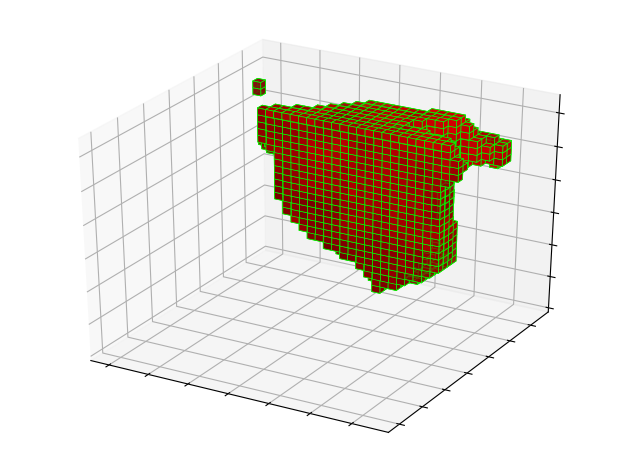}}\\%
        
        \rotatebox[origin=c]{90}{\scriptsize Labeled 20\%}&%
        \fcolorbox{lime}{white}{\includegraphics[align=c,width=\linewidth,height=1.3cm]{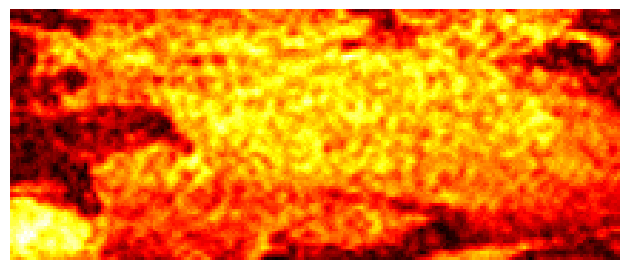}}&%
        \fcolorbox{magenta}{white}{\includegraphics[align=c,width=\linewidth,height=1.3cm]{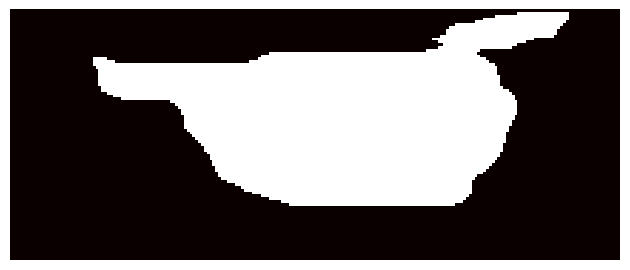}}&%
        \fcolorbox{cyan}{white}{\includegraphics[align=c,width=\linewidth,height=1.3cm]{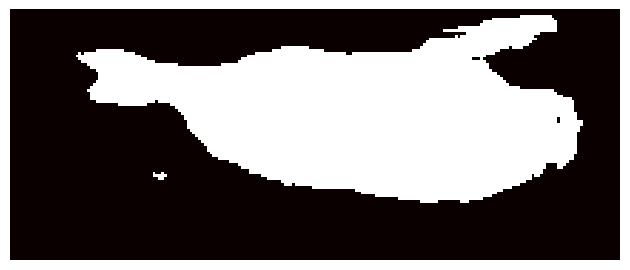}}&%
        \fcolorbox{lime}{white}{\includegraphics[align=c,width=\linewidth,height=1.3cm]{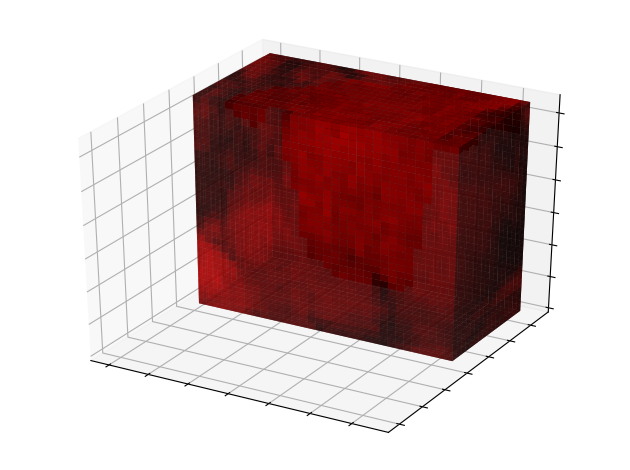}}&%
        \fcolorbox{magenta}{white}{\includegraphics[align=c,width=\linewidth,height=1.3cm]{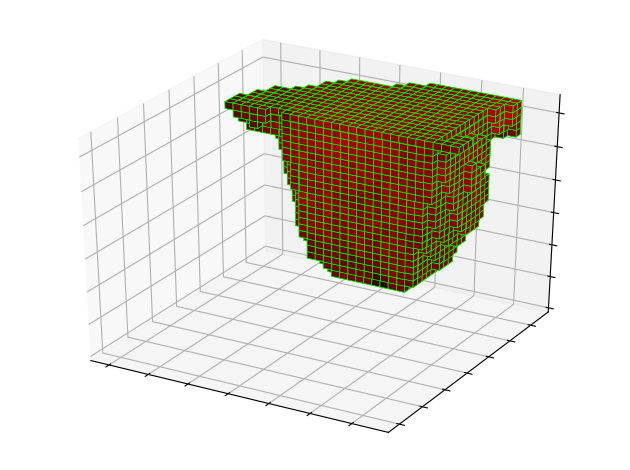}}&%
        \fcolorbox{cyan}{white}{\includegraphics[align=c,width=\linewidth,height=1.3cm]{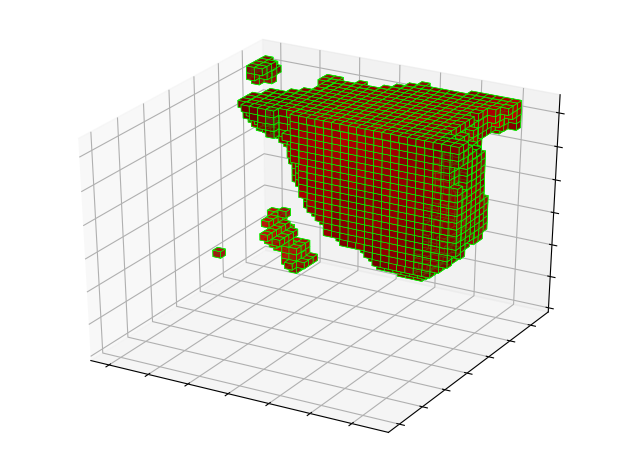}}\\%
        
    \end{tabularx}
    \caption{Qualitative results with 3D images on the LA dataset~\cite{XIONG2021101832}. We provide two views for better readability -- a slide view and a 3D view.}
    \label{fig:atrial-qualitatite-results}    
\end{figure*}

\subsection{Ablation Studies}
\label{sec:sec:ablation}

\paragraph{Softmax temperature hyperparameter} We perform an ablation study varying the temperature hyperparameter defined in Eq.~\ref{eq:swta}. We show results concerning SWTA-TSA over the three datasets with the regime at 20\% in Tab.~\ref{tab:ablation_temp}. We observe that, concerning the GlaS dataset, increasing temperature from 1 to 100 has a positive impact on performance until a plateau is reached at that point; instead, we obtained the best performance with a temperature of 20 for the PH2 and HMEPS datasets.

\paragraph{Hebbian variants} Tab.~\ref{tab:ablation_hebb_formulations} concisely ablates the results obtained by exploiting the different Hebbian learning strategies introduced in Sec.~\ref{sec:sec:method_hebbian_tconv}. Specifically, we report the outcomes in terms of the Dice Coefficient for the GlaS dataset (the most popular and challenging among those used)
%three datasets 
considering different percentages of label availability. In 
%almost 
all cases, our SWTA-TSA Hebbian unsupervised learning formulation achieves the best performance, motivating its selection as our preferred choice.
%; consequently, in the paper, we reported the results concerning the latter for all the experiments.
%, motivating SWTA-TSA as our preferred choice. 

\paragraph{Hebbian unsupervised first stage} We conduct an ablation study to assess the Hebbian unsupervised pre-training independently. Specifically, following the literature, we exploit linear probing, where a linear classifier is trained on top of the frozen learned features~\cite{DBLP:conf/iccv/GansbekeVGG21,DBLP:conf/iccv/JiVH19}. Tab.~\ref{tab:ablation_first_stage} presents the Dice Coefficient results on the GlaS dataset, comparing our approach with two other unsupervised pre-training methods. Our approach achieves the best performance, demonstrating its superiority.

\subsection{Experiments with Volumetric Images}
\label{sec:sec:3d-exp}

\noindent To prove that our approach can be easily extended to 3D medical images, we report some outcomes using the Left Atrial (LA) dataset~\cite{XIONG2021101832}. Specifically, it contains 100 3D MRI images from the 2018 atrial segmentation challenge; following the literature, we use 80 images for training and 20 for testing.

The results of our evaluation are shown in Tab.~\ref{tab:comparison-sota-la}. Our approach ranks best or second best most of the time, and our Hebbian initialization yields benefits to one-stage SOTA methods. Qualitative results can instead be found in Fig.~\ref{fig:atrial-qualitatite-results}
%, among the competing methods, and it confirms to be particularly promising in those scenarios with more scarce label availability (up to 10\%).

\begin{figure*}[!tb]
    \centering
    \begin{subfigure}[t]{0.48\textwidth}
        \center\includegraphics[width=0.7\textwidth]{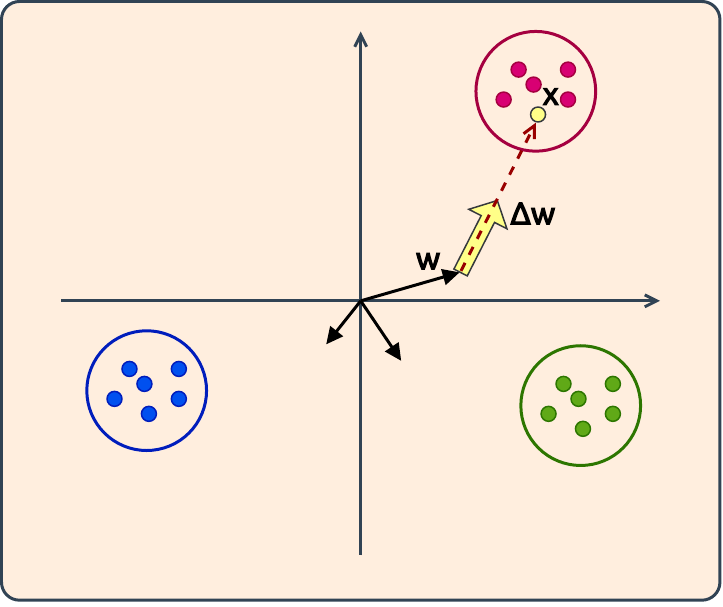}
        \caption{\footnotesize 
        Starting from a random initialization of weight vectors, inputs are presented to the neurons. The weight vectors take an update step towards the input position in the data plane. %Neurons take steps of longer relative length towards input data points to which they are closer.
        %Weight vectors are initialized randomly. During learning, neurons update their weight vectors toward the currently observed data point. The sizes of these steps depend on how close the weight vectors are to the inputs; in the end, each neuron tends to specialize on a different set of data.
        }
        \label{fig:swta_a}
    \end{subfigure}
    ~
    \begin{subfigure}[t]{0.48\textwidth}
        \center\includegraphics[width=0.7\textwidth]{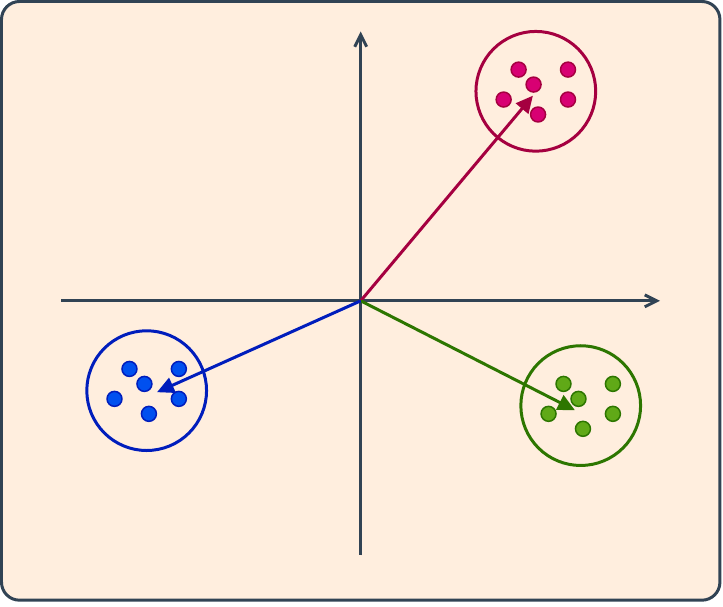}
        \caption{\footnotesize 
        After multiple iterations, different weight vectors converged to different cluster centroids, thanks to the SWTA mechanism.
        %Final configuration after convergence. Thanks to the SWTA mechanism, different neurons have specialized in different data types. Therefore, each neuron has learned to represent the centroid of a particular cluster in the data space.
        }
        \label{fig:swta_b}
    \end{subfigure}
    %\caption{Hebbian updates following SWTA-type learning.}
    \caption{ 
    The illustration represents weight vectors (arrows) and data points (dots) within a data space. Fig.~\ref{fig:swta_a} shows how a weight vector updates according to the SWTA equation when input is introduced. Fig.~\ref{fig:swta_b} depicts the final positions of the weight vectors after multiple iterations, showing that each has converged to a distinct cluster centroid, guided by the SWTA mechanism.}
    \label{fig:swta}
\end{figure*}

%%%%%%%%%%%%%%%%%%%%%%%%%%%%%%%%%%%%%%%%%%%%%%%%%%%%%%%%%%%%%%%%%%%%%%%%%%%%%%%
\section{Conclusion}
\label{sec:conclusion}
\noindent In this work, we presented a novel two-stage semi-supervised approach for semantic segmentation, leveraging bio-inspired learning models for a first unsupervised pre-training step, followed by a second supervised fine-tuning phase. Our core contribution is the formulation of Hebbian learning rules for transpose-convolutional layers, constituting the upsampling path of many popular semantic segmentation architectures. Experiments over several biomedical image segmentation benchmarks using different degrees of labeled data demonstrated the effectiveness of our methodology compared to other SOTA approaches. Furthermore, we also explored combinations of our Hebbian pre-training approaches with existing pseudo-labeling and consistency-based semi-supervised methods, resulting in performance improvements.
These findings hold significant practical relevance, especially considering the extreme cost required for collecting annotated data, particularly in the biomedical domain, and because bio-inspired algorithms can be advantageous for developing biologically plausible models.

%The proposed methodology still presents some limitations, which can be addressed in future works. One such limitation is the lack of a theoretical and experimental exploration of Hebbian approaches with 3D/volumetric data, which is frequently encountered in biomedical scenarios. 

%\section*{Acknowledgments}
%This should be a simple paragraph before the References to thank those individuals and institutions who have supported your work on this article.

\appendix

\section{Hebbian Learning Background}
\label{appendix:background}

The methodology outlined in our work draws inspiration from the biological processes of synaptic adaptation and learning. While Hebbian theory has its roots in neurobiology, a background in neuroscience is not required to grasp the principles underlying these methodologies. Interestingly, mathematical models of Hebbian learning reveal surprising parallels with well-known machine learning concepts, such as clustering, Principal Component Analysis (PCA), and other mechanisms for unsupervised feature extraction. In the following sections, we provide additional context to enhance the reader's understanding of how certain learning principles emerge from the proposed learning rules. Specifically, we introduced two learning principles: Soft-Winner-Takes-All (SWTA) and Hebbian Principal Component Analysis (HPCA). We begin by examining SWTA in greater depth, highlighting its connections to standard centroid-based clustering. Subsequently, we delve into HPCA, demonstrating how this synaptic model facilitates the identification of principal components from data.

%In this section, we provide a more detailed background about Hebbian biologically inspired models of synaptic plasticity. We already pointed out that competitive learning rules such as Soft-Winner-Takes-All (SWTA) enable the extraction of cluster centroids by neurons, while Hebbian Principal Component Analysis (HPCA) allows the extraction of data principal components. Here, we report further details explaining these two techniques.

\paragraph{Soft-Winner-Takes-All (SWTA).}
The SWTA weights update rule expressed in Eq.~\ref{eq:swta} is reported again, for convenience, hereafter:
%For simplicity, we report again the learning equations for SWTA that we already defined in the paper:

\begin{equation} 
    \label{eq:swta_rep}
    \Delta w_{i, j} = \eta \, \mathrm{softmax}(y_1, y_2, ...)_j \, ( x_i - w_{i, j} ) ,
\end{equation}

\noindent For a given neuron $i$, we can distinguish two multiplicative components. The first component includes the softmax, together with the learning rate. It is a scalar (one term for each neuron) that essentially modulates the length of the weight update step for the given neuron. The second component is $x_i - w_{i, j}$, which is instead a vector. This is the direction of the weight update. Intuitively, the neuron takes a small update step in the direction that links its weight vector $w_{i, j}$ with the input $x_i$. When this process is repeated over and over for many inputs, the weight vector eventually converges to the centroid of the observed inputs (Fig.~\ref{fig:swta}). Furthermore, since the softmax operation assigns modulation coefficients close to 1 for highly active neurons and near 0 for less active ones, this mechanism enables different neurons to specialize in distinct input clusters, as neurons that take a larger step toward a specific input are more likely to produce stronger responses when similar inputs are encountered in the future.

%\noindent where the notation follows the one used in the paper. The softmax operation can be computed using a temperature hyperparameter $t$, whose best value depends on the specific task and should be empirically optimized. 
%From~\cref{eq:swta_rep}, we can notice that the weight vector of a neuron $j$ is updated along the direction $x_i - w_{i, j}$, where $i=(1, 2, \dots)$. This direction links the weight vector $w_{i, j}$ to the current input $x_i$ along a straight line. 
%The softmax and the learning rate, instead, represent a multiplicative factor that modulates the size of this update.
%Therefore, the weight vector is updated by taking a small step toward the current data point observed by the neuron. As shown in~\cref{fig:swta}, different neurons can specialize in different data clusters thanks to this mechanism.

\paragraph{Hebbian Principal Component Analysis (HPCA).}
For convenience, we report again Eq.~\ref{eq:hpca} concerning the HPCA weights update hereafter:
%As before, the HPCA update rule from the main paper is also reported, for convenience:
%Again, for simplicity, we report the learning equation for HPCA defined in the paper:

\begin{equation} 
    \label{eq:hpca_rep}
    \Delta w_{i, j} = \eta \, y_j \, ( x_i - \sum_{k=1}^j y_k w_{i, k} ) .
\end{equation}

%\noindent where the symbols have the same meaning as in the main paper. 
%\noindent Again, the notation follows the one defined in the paper. Similarly to the previous case about SWTA, we can conduct an analysis of~\cref{eq:hpca_rep} starting from a single linear neuron:

\noindent The learning process underlying this rule is less intuitive to visualize compared to the SWTA case, but valuable insights can be gained by examining the conditions required for convergence to a stochastic equilibrium. Specifically, for this equilibrium to be achieved, the average weight update across all inputs must equal zero. Mathematically, this can be written as:

\begin{equation} 
    \mathbb{E}[\Delta w_{i, 1}] = \mathbb{E}[\eta \, y_1 \, ( x_i - y_1 w_{i, 1} )] = 0 ,
\end{equation}

\noindent where the equation refers to neuron 1 in particular, but the following arguments apply also to any other neuron. By rearranging the terms, we obtain:
%\noindent from which we obtain:

\begin{equation} 
   \mathbb{E}[ y_1  x_i ] = \mathbb{E}[ y_1 y_1 w_{i, 1} ] .
\end{equation}

\noindent At this point we can recall the input-output relationship of the neuron, i.e., $y_1 = \sum_k x_k w_{k, 1}$, and substitute it in the previous equation as follows: 
%\noindent Let's rewrite $y_1$ in terms of $x_i$ and $w_{i, 1}$, \ie $y_1 = \sum_k x_k w_{k, 1}$:

\begin{equation} 
   \mathbb{E}[ \sum_k x_k w_{k, 1}  x_i ] = \mathbb{E}[ \sum_k x_k w_{k, 1} \sum_h x_h w_{h, 1} w_{i, 1} ] .
\end{equation}

\noindent Since the sum can commute with the expectation, and since $w_{i, j}$ does not depend on $x_i$, after rearranging the terms we can derive the following equation:
%\noindent Rearranging the terms, commuting the expectation and the sum, and noting again that $w_{i, j}$ does not depend on $x_i$, we obtain:

\begin{equation} 
  \sum_k  w_{k, 1} \mathbb{E}[ x_k x_i ] = \sum_{k, h} w_{k, 1} \mathbb{E}[ x_k x_h ] w_{h, 1} w_{i, 1} .
\end{equation}

\noindent It can be noticed that $\mathbb{E}[ x_i x_k ]$ is the input data covariance matrix, while $\sum_{k, h} w_{k, 1} \mathbb{E}[ x_k x_h ] w_{h, 1}$ is a scalar. By calling the first term as $C_{i, k}$ and the second term as $\lambda$, we can simplify the equation to:
%It can be noticed that $\mathbb{E}[ x_i x_k ]$ is the input data covariance matrix, call it  $C_{i, k}$, while $\sum_{k, h} w_{k, 1} \mathbb{E}[ x_k x_h ] w_{h, 1}$ is a scalar, say $\lambda$. This simplifies the equation to:
%\noindent At this point, the term $\mathbb{E}[ x_i x_k ]$ is the covariance matrix $C_{i, k}$ of variables $x_i$, $i=(1, 2, \dots)$, and the quantity $\sum_{k, h} w_{k, 1} \mathbb{E}[ x_k x_h ] w_{h, 1}$ is just a constant, which we can call $\lambda$. Therefore, the resulting equation

\begin{equation} 
  \sum_k  w_{k, 1} C_{i, k} = \lambda w_{i, 1}.
\end{equation}

\noindent This is a familiar equation of the eigenvalues and eigenvectors of the matrix $C_{i, k}$. In other words, this tells us that, according to the synaptic dynamics defined in Eq.~\ref{eq:hpca_rep}, equilibrium of neuron $1$ is achieved when the weight vector $w_{i, 1}$ converges to an eigenvector of the covariance matrix (and $\lambda$ is the corresponding eigenvalue). This is exactly the definition of a principal component.

\section{Implementation Details}
\label{appendix:implementation}
In this section, we provide some implementation details. We also refer the reader to our repository available at \url{https://github.com/ciampluca/hebbian-bootstraping-semi-supervised-medical-imaging}.

Our model is implemented using PyTorch, with all training and inference processes conducted on an NVIDIA DGX-A100. To ensure fair comparisons, we maintained consistent settings across all experiments. For the 2D datasets, data augmentation during training includes vertical and horizontal flips as well as rotations. Input images are resized to $128 \times 128$ for both training and inference. In contrast, for the LA dataset~\cite{XIONG2021101832}, the same augmentation strategy is applied during training. Random patches of size $96 \times 96 \times 80$ are extracted during training, while inference employs a sliding window approach with a 0.5 overlap ratio, using patches of the same size.

For the 2D datasets, we use the SGD optimizer along with a multi-step learning rate scheduler to train all our models. The initial learning rate is set to 0.5, and it is reduced by a factor of 10 every 50 epochs. In contrast, for the 3D LA dataset, the initial learning rate is set to 0.1. For some of the competitor techniques used in comparison, we opted for a lower initial learning rate, as we observed better results empirically. Specifically, we set the learning rate to 0.001 for the unsupervised phase of the VAE and SSL-based pipelines for the 2D datasets, and 0.0001 for the LA dataset. Additionally, for the weight $\lambda$ in the unsupervised loss functions of the competitor pseudo-labeling/consistency-based methods, we increase $\lambda$ linearly with the number of epochs following previous works, i.e., $\lambda = \lambda_{max} \times \frac{epoch}{max\_epoch}$, where $\lambda_{max}$ is set to 5. 

We set the total number of epochs to 200. In the first unsupervised stage, we always get the model snapshots obtained from the last training epoch to initialize the models of the second fine-tuning step. In fact, Hebbian-learned models usually converge to a stable configuration of weights that do not change significantly after a few epochs of training. We empirically set the end of the training to 200 epochs as we noticed that all models' parameters on all datasets were not changing significantly after that point. Then, for the final evaluation of the test set, we pick the best model in terms of DC on the validation split obtained during the fine-tuning stage.

To ensure that our results are statistically relevant, we implement a 10-fold cross-validation protocol (5-fold for the LA dataset), varying the training and test splits.

\bibliographystyle{elsarticle-num} 
\bibliography{biblio}

\end{document}